A COMPUTATIONAL MODEL OF
SERIAL AND PARALLEL PROCESSING IN VISUAL SEARCH

BY

RACHEL FLOOD HEATON

DISSERTATION

Submitted in partial fulfillment of the requirements
for the degree of Doctor of Philosophy in Psychology
in the Graduate College of the
University of Illinois Urbana-Champaign, 2023

Urbana, Illinois

Committee members:

    Associate Professor Emeritus Robert Wickesberg, Chair and Co-Director
    Leonidas A. A. Doumas, Senior Lecturer, University of Edinburgh, Co-Director
    Professor Diane Beck
    Benjamin D. Evans, Lecturer, University of Sussex
    Associate Professor Jonathan Livengood
    Professor Ranxiao Frances Wang
    Professor John Hummel

## Abstract

The following is a dissertation aimed at understanding what the various phenomena in visual search teach us about the nature of human ventral visual representations and processes. I first review some of the major empirical findings in the study of visual search. I next present a theory of visual search in terms of what I believe these findings suggest about the representations and processes underlying ventral visual processing. These principles are instantiated in a computational model called CASPER (Concurrent Attention: Serial and Parallel Evaluation with Relations), originally developed by Hummel that I have adapted to account for a range of phenomena in visual search. I next present simulations using CASPER to account for some of the major empirical findings in the literature on visual search. I then describe an extension of the CASPER model to account for our ability to search for visual items defined not simply by the features composing those items but by the spatial relations among those features. Seven experiments (four main experiments and three replications) are described that test CASPER's predictions about relational search. Finally, I evaluate the fit between CASPER's predictions and the empirical findings and show with three additional simulations that CASPER can account for negative acceleration in search functions for relational stimuli if one postulates that the visual system is leveraging an emergent feature that bypasses relational processing.

## Acknowledgements

This work would not have been possible without the support of many people. The following words are inadequate to capture the depth of my gratitude to them. I couldn't have finished this work without Robert Wickesberg's generous service and support as the chair of my committee. I fear I will never be able to repay him for his kindness and steadiness, but I will try my best to pay it forward. I believe that if we all acted according to Bob's example, everything would be okay.

Thank you to my committee. John Hummel taught me everything I know about computational modeling. He inspired me to greater ambition and creativity in my research. I also credit him with reinforcing for me the importance of honesty in how I approach modeling. ("We don't just report what works. We also report the ways it doesn't work!") Diane Beck is an amazing researcher and educator who as a matter of scheduling circumstances taught me the majority of what I know about the vision literature. Her seminars provided the foundation for my understanding of my work, and I can still hear her voice in my head when I think about certain studies. I also thank her for her suggestions to improve the work in this dissertation.

John Livengood somehow managed to convince me over the years that philosophy is important to the way I think about my research. This is an amazing accomplishment. I deeply value the conversations we have had, and I thank him for his trustworthiness and generosity. Alex Doumas has been a constant positive example of a principled modeler and scientist. I always look forward to opportunities to talk with Alex about science, and he has given me great advice when I needed it. Ben Evans has always been a reliable source of sharp insights and sharp wit, and I thank him for his suggestions that helped improve this work. Frances Wang has always been a steady presence who asks great questions. I learned a great deal from watching her teach and I appreciate her willingness to serve on my committee.

I would also like to thank the rest of the current and former members of the A&P program area and the broader Department of Psychology, including faculty Dan Simons, Simona Buetti, Alejandro Lleras, and David Irwin. I learned a great deal from all of them, especially from conversations with Simona Buetti and Alejandro Lleras about issues in visual search. Thank you also to Drs Buetti and Lleras for helpful conversations about the experiments reported here. Thank you also to the other graduate students in the department, especially Zoe Xu and Gavin Ng for developing lab code that I adapted for the online experiments reported here.

Thank you to Firmino Pinto for outstanding technical support within the Department of Psychology. I would additionally like to thank June Clark Eubanks, Ashley Ramm, Sarah Challand, Lori Butler, Brenda Reinhold, Melissa Odom, and the rest of the department staff for deftly dispatching with red tape during my PhD—you are amazing at what you do, and you made this process much easier than it could have been.

I gained a great deal of knowledge from my interactions with people who opened their lab meetings to me at various times, including Aaron Benjamin, Jon Willits, Jeff Bowers, Keith Holyoak, Hongjing Lu, and Jeremy Wolfe. Thank you also to Margarita Pavlova and Matthew Slocombe for organizing the Analogical Minds seminars, which were extremely valuable. I also am grateful to Cliff Shin, Gerry Guthrie, and Matthew Peterson for setting me on a path toward cognitive science while I was studying design.

The work presented here was financially supported by the National Science Foundation Grant BCS1921735 and completed with funding from the Department of Psychology at the University of Illinois. I would also like to thank organizations outside the department who gave me opportunities and funding during my doctoral work, including the Siebel Center for Design and the New Frontiers Initiative at Illinois.

On a more personal note, I would like to thank every person who chooses kindness in the way they approach the world. I would especially like to thank Brittney Currie for her undying support, wisdom, optimism, resilience, and amazing sense of humor. I would also like to thank Kerrith Livengood for her advice and friendship while I was working on this degree.

Thank you to my parents, particularly my mother, for giving me an amazing range of experiences as a child which I believe helped me succeed academically. Thank you also to my mom and Bryan Heaton for their flexibility and their generous help with parenting when I needed it, which was often.

Finally, thank you to my beautiful daughter, Vala. During the darkest times you were what kept me going. I love you.

# Table of Contents

# Chapter 1: Visual search as a window into visual representation and processes

Visual search is among the most studied tasks in perception and cognition. In addition to its direct relevance as a task people perform in the "real world", visual search has long been viewed as a window into the nature of human visual representations and processes. Understanding the psychological representations and processes involved in visual search amounts to understanding much of vision, including visual representations and processes in the ventral processing stream (i.e., leading from V1 to V2, V4, lateral occipital complex (LOC), and inferotemporal cortex) as well as processes in dorsal vision (Goodale et al., 1991; Tanaka, 1996).

In a visual search experiment, research participants are typically asked to search for a *target* item among a field of *distractor* items. There are numerous variations on this basic task: sometimes the exact target is unknown (e.g., Bravo & Nakayama, 1992; Müller et al., 1995). Alternatively, there may be no target present, or even multiple targets present (e.g., Townsend, 1990; Thornton & Gilden, 2007). Visual search has been so well explored, with so many variations, that there have been grumblings that visual search experiments have become more concerned with studying a paradigm than studying the visual system.

But visual search endures as a popular and fruitful paradigm for studying vision because it has several attractive properties that allow for direct insights into human visual representations and processes. First, visual search can be performed without verbal responses on the part of the participant. In this way, the experimenter can be fairly confident that they are for the most part characterizing visual representations rather than, for example, linguistic processes. Contrast this with a paradigm like an object naming task, where care must be taken to determine whether an effect is due to visual-, conceptual-, or language-based processes (see Biederman & Cooper, 1991a).

Visual search also allows the experimenter to manipulate a number of factors that are important for any visual task, not just search for an item. These include the number of items to which a person might potentially attend, the features of those items, and their locations in the visual field (e.g., how close they are to each other, or how far they are located in the periphery). Manipulating these factors can dramatically change the difficulty of a task. Some displays can be searched, in effect, all at once, while others require a great deal of effort.

Whether items in a display or scene are processed one at a time under attention, in parallel all at once, or some combination thereof, matters for all kinds of human behaviors. Among these are object recognition, similarity judgments, scene perception, and even real-world tasks such as finding a setting in a software menu or turning on an appliance. If we understand the combinations of conditions that give rise to efficient parallel processing vs. slow serial processing, we can better predict the way people will interact with the visual world.

**A roadmap for this document**

The following is a dissertation aimed at understanding what the various phenomena in visual search can teach us about the nature of human ventral vision. I will first review some of the major empirical findings in the study of visual search over the last four decades. I will next present the broad strokes of a theory of visual search in terms of what I believe these findings suggest about the representations and processes underlying ventral visual processing. These principles are instantiated in a computational model called CASPER (Concurrent Attention: Serial and Parallel Evaluation with Relations). An unpublished proof of concept of this model was first implemented by Hummel in 2018. That proof-of-concept model predicted qualitative differences between feature and conjunction searches and captured the negatively-accelerating search functions characterizing some visual search paradigms. I have subsequently adapted the model's mechanisms to account for the wider range of phenomena in visual search that I will report here.[1] I will describe the details of CASPER's operation as they stand today and their relation to my broader theoretical claims. I will next present simulations using CASPER to account for some of the major empirical findings in the literature on visual search. I will then describe an extension of the CASPER model to account for our ability to search for visual items defined not simply by the features composing those items but by the spatial relations among those features. I will also present seven experiments (four main experiments and three replications) testing CASPER's predictions about relational search. Finally, I evaluate the fit between CASPER's predictions and the empirical findings and show with three additional simulations that CASPER can account for negative acceleration in search functions for relational stimuli if one postulates that the visual system is leveraging an emergent feature that bypasses relational processing.

---

[1] One commonality between the original model by Hummel and the model presented in this dissertation is the use of concurrent parallel and serial attention processes, but nearly all of the other details of its operation have been changed.

**Major Empirical Phenomena in Visual Search**

The empirical literature on visual search is enormous, but there are several significant findings for which any model of visual search should account. These results can be organized according to the insights they provide into either visual processes or representations.

In the domain of visual processing, a major concern in the visual search literature has been the degree to which the processing of items in a display proceeds in parallel across the items, versus serially one item at a time. As elaborated below, Treisman & Gelade (1980) provided evidence that search for a target defined by a single feature (e.g., color) can proceed in parallel over all search items, while search for a target defined by a conjunction of features (e.g., color and shape) requires focused attention on one (or at most a few) item(s) at a time. However, Wolfe et al. (1989) later showed that some conjunction searches benefit from parallel processing so much that they are nearly as parallel as some feature searches. Buetti et al. (2016) showed that feature search, too, exhibits varying degrees of parallel versus serial processing, depending on the similarity of the search target to the distractor items, consistent with Duncan & Humphreys' (1989) theory of the effect of target-distractor similarity on search difficulty.

Other findings tell us about the representations the visual system uses during search. Treisman & Souther (1985) demonstrated that when a target item contains a superset of the low-level features of the distractors, search is easier than when the target only contains a subset of the features of the distractor, indicating that the mental representation of the search target can differentially influence search difficulty. Pomerantz et al. (1977) showed that the configuration of line segments into meaningful higher-order shapes can also facilitate search.[2] However, from Logan (1994), we know that searches that require subjects to represent and compare spatial relations are extremely difficult.

*Binding via attention*

One early and influential theory of visual search, Treisman & Gelade's (1980) *Feature Integration Theory* (FIT), postulated that a major function of visual attention is to bind independent visual features (such as colors and shapes) into coherent items. Their 1980 paper presented nine experiments testing and supporting this hypothesis. Among their major findings was that visual search for a target (e.g., either a blue X or a brown or green S) that can be distinguished from non-targets (e.g., brown Ts and green Xs) by a single feature proceeds largely in parallel (i.e., search response times were largely independent of the number of distractors in the display), whereas search for a target (e.g., a green T) that can only be distinguished from the same distractor items based on a conjunction of multiple features proceeds serially (i.e., search times increase linearly with the number of distractors items in the display). Treisman and Gelade also presented several other findings supporting their hypothesis that feature binding is a central function of visual attention.

---

[2] See also Treisman & Souther (1985) Experiment 4

*Efficient conjunction search: guiding attention via parallel processing*

Wolfe et al. (1989) observed that a limitation of FIT is that, according to that theory, beyond simply activating the basic features of search items, parallel processing does nothing to guide the selection of search items for attentional scrutiny. Wolfe et al. (1989) showed that not all searches for conjunctions result in steep linear search functions. In their experiments, conjunction searches produced search functions that negatively accelerate as a function of set size, suggesting that some searches that cannot be performed entirely in parallel nonetheless still benefit from some parallel processing.

In the original[3] *Guided Search* (GS) model, Wolfe et al. (1989) postulated that parallel and serial processing are not autonomous (i.e., such that parallel processing

---

[3] Starting with Guided Search 2.0, Wolfe (1994) placed less emphasis on continuous parallel processing informing a strictly serial process than in the original theory. Guided Search 4.0 subsequently provided a more specific account of attentional processing

finishes before serial processing begins) but that parallel processing continuously informs, or guides, serial selection. Under this scheme, parallel processing of relevant features activates search items and the most active item in the display is chosen for serial attentional selection. In the original version of GS, emphasis was placed on the continuous interplay of parallel guidance with serial selection, where the target is increasingly likely to be attended as parallel processing provides guidance.

*Target-distractor similarity and search efficiency*

Wolfe et al. showed that parallel processing influences conjunction searches (counter to the predictions of FIT), and Buetti et al. (2016) showed that the effect of parallel processing also varies for feature-based searches. Using a constant search target across conditions, Buetti et al. manipulated target-distractor similarity by varying both the shape and color of the distractors. For a search target defined as a red triangle, search was more difficult (i.e., the slope of the function was steeper) when the distractors were orange diamonds than when they were blue circles. This result is consistent with Duncan & Humphreys' (1989) account of the search surface defined by target-distractor and distractor-distractor similarity, which predicts that high target-

---

where search items enter visual working memory one at a time, such that multiple items in visual working memory are subjected to an asynchronous attentional diffusion process (Wolfe & Gray, 2007).

distractor similarity will result in a difficult search, while low target-distractor similarity will result in an easier search.

By systematically manipulating target-distractor similarity, Buetti et al. (2016) found a range of difficulty in feature searches. However, in all conditions, the functions of RT by set size were negatively accelerating, more consistent with a log function than a linear function. Buetti et al.'s negatively accelerating functions for feature searches, along with Wolfe et al.'s (1989) negatively accelerating functions for conjunction searches, demonstrate that multiple factors contribute to the steepness of the search function that cannot be characterized simply in terms of the dichotomy between feature vs. conjunction search. Instead, both feature and conjunction searches can be more or less efficient —that is, they can show a greater or smaller contribution of parallel processing—in terms of how they proceed. An efficient search will benefit more from parallel processing, while an inefficient search will not derive as much benefit from parallel processing.[4]

*Search asymmetries and the target template*

One phenomenon in visual search that helps us to understand parallel versus serial processing is search asymmetries. Search asymmetries occur when reversing the characteristics of the target item with the distractors results in a different level of difficulty in finding the target. For example, Treisman & Souther (1985) in their Experiment 1 demonstrated that a search for a Q-like shape among Os is substantially easier than a search for an O among Qs. Treisman and Souther characterized this effect in terms of the need to bind features: If a distinguishing feature in the target is present in only one location in the visual field, it will be detectable via parallel processing. If there is no such distinguishing feature, then scrutiny is required.

---

[4] Kristjánsson (2015) also makes the point that there is not a clear dichotomy between steep serial and flat parallel search functions in visual search.

A more general way to characterize this finding is to say that the presence of a visual feature in a search item is more salient than its absence. As elaborated later, such asymmetries also suggest that evaluating whether a search item is or is not the search target is based on a similarity measure more complex than, say, a Euclidean distance, a simple dot product, or a cosine measure of vector similarity, all of which are standard measures of vector similarity but are also symmetric, meaning they are not sensitive to the difference between the presence or absence of mismatching features.

*Configural effects*

Representational hierarchy also appears to be a factor in visual search efficiency. In several experiments, Pomerantz et al. (1977) showed an effect of representation level on search difficulty. In their experiments, search items were defined over collections of low-level features (i.e., arrangements of oriented line segments). Pomerantz et al. manipulated the configural context of the line segments. In a baseline condition, the target was a simple 45-degree oriented line among orthogonally oriented distractors. In a good configural context condition, extra line segments were added to the baseline stimuli such that new meaningful shapes were created (i.e., triangles and arrows) that could help differentiate the target from the distractors. In a poor configural context condition, the extra line segments did not combine into any sort of meaningful higher-order shape. Search was substantially more efficient in the good configural context condition than in the baseline and poor configural context conditions. Moreover, this effect occurred regardless of crowding caused by the added features compared to the baseline condition, which should make the task harder.

Pomerantz et al.'s results suggest that higher-order representations have a proportionally greater influence on search efficiency than lower-order representations. Pomerantz et al.'s findings are compatible with Palmer's (1977) result showing that good configuration of line segments leads to faster timed verification of parts within visual stimuli, and various other studies that show preferential processing for arrangements of visual stimuli that are consistent with longer contours, axes of symmetry, surfaces, and parts (Pashler, 1990; Wenderoth, 1994; Nakayama et al., 1995; Elder & Zucker, 1993; Kovacs & Julesz, 1993; Singh & Hoffman, 2001).

*Search for relations*

Relational processing requires greater attentional demands than feature-based processing (Markman & Gentner, 1993; Stankiewicz et al., 1998; Tohill & Holyoak, 2001; Thoma & Davidoff, 2002; Thoma et al., 2007). Relational representations can be visible (e.g., one object above another object) or abstract (e.g., objects in a functional relationship with each other; Green & Hummel, 2006). The binding of relational roles to their fillers (e.g., in the *above* (X, O) relation, binding the *above* role to X and the *below* role to O) is thought to happen late in ventral visual processing (i.e., LOC or later; Kim & Biederman, 2011; Christoff et al., 2001; Kroger et al., 2002), at a point where perception and cognition are not so easily distinguished from each other.

Whereas a conjunction search task requires dynamic binding of features that are present at a location in the visual field into an object in order to detect a search target, a relational search task further requires that relational roles are bound to their fillers such that they are not confused with each other. For example, an item that contains a red circle above a blue square would not be confused with a blue circle above a red square, a blue square above a red circle, or a pair of purple circle-squares. Relational binding therefore requires a more complex account of dynamic binding than the one offered by FIT: FIT assumes that attention (and therefore binding) is limited to one object at a time, precluding the binding of two (or more) objects simultaneously to their relational roles.

Accordingly, Logan (1994) showed that a search for a target defined by a relation between two objects (e.g., a + above a - among distractors depicting - above +) appears to demand a great deal of attention. Search over these relations in Logan's experiments was inefficient and produced a steep linear search slope of about 85ms/item. This result is consistent with an inefficient search that requires serial deployment of attention without benefit from parallel processing. Notably, Logan obtained a search slope in relation-based searches that was approximately 3x the search slope obtained in conjunction search by Treisman & Gelade (1980). Logan also found that participants did not improve in search efficiency with practice. Hence relational search appears to be one of the most demanding tasks in the domain of visual search.

**Models of visual search**

Numerous theories have been proposed to explain the phenomena in visual search. Some of these theories are computationally instantiated, while others are verbal theories. Some influential verbal theories include Treisman & Gelade's (1980) FIT, which explored the roles of parallel processing and serial attentional binding in search, and Duncan & Humphreys (1989) account of the effects of target-distractor and distractor-distractor similarity on search efficiency. Various incarnations of Wolfe's Guided Search models, (Wolfe et al.,1989; Wolfe, 1999; Wolfe & Gray, 2007; Wolfe, 2021) have hypothesized a role for parallel processing in guiding attention to a search target, and some basic simulations of these theories provide evidence for plausibility for the account. Bundesen's Theory of Visual Attention (TVA; 1990, 1998) provided an account of the role of feature weighting in attentional selection, proposing the use of Luce's (1959) choice axiom rather than random or deterministic attentional selection. Logan's Code TVA (1990) extended Bundesen's selection mechanism to include an account of the influence of grouping by proximity on attention. Lleras et al. (2020) introduced Target Contrast Signal Theory (TCST) which provided an account of parallel processing during search that was based on differences between search items.

Of the theories mentioned above, the ones that are computationally instantiated aim to provide a mathematical characterization of search functions using abstracted approximations of differences between a target and distractor items (i.e., all the differences in properties between search items might be embodied in the calculations as a single scalar value). Marr (1982) would likely have categorized these models as *computational theory level* models. To the best of my knowledge, no prior models have attempted to provide a detailed process (i.e., *algorithmic level*) model that specifies how the visual representation of targets and distractors gives rise to the observed search behavior. One of the goals of the work described here is to provide such an account.

CASPER was initially developed as a response to Lleras et al.'s (2020) TCST. TCST proposes that a parallel stage of processing accumulates information about contrast between the target template and search items. Items that accumulate enough contrast from the target are removed from contention during parallel processing. TCST then assumes a timeout period must elapse between the last item rejected via parallel processing and the beginning of the serial phase of search. In TCTS, the serial selection is random, and items are selected until the target is found or all are rejected. Lleras et al. showed that TCST can properly mathematically characterize a range of difficulties for feature searches like those reported in Buetti et al. (2016). However, the inclusion of the timeout after parallel processing and before serial processing in TCST, although mathematically convenient to predict search functions, may be problematic for TCST as a process model. Particularly from an evolutionary perspective, a wait between parallel and serial processing might be a big disadvantage. There are a number of scenarios where an organism's survival is unlikely if parallel processing must strictly precede serial processing (e.g., when the organism is either predator or prey).

      The CASPER model builds on the idea of contrast accumulation from TCST and also incorporates insights from other theories of visual search in the prior literature. The result is a computational process model that can account for a range of visual search results.

# Chapter 2: The CASPER Model

## Core Theoretical Claims

The CASPER model embodies several core theoretical claims, summarized in Table 2.1. One of the most important is that visual search does not follow a strict two-stage process of the kind that is hypothesized in many models of visual search (e.g., Hoffman, 1979; Treisman & Gelade, 1980; Lleras et al., 2020), where parallel processing concludes before serial processing begins. Instead, CASPER assumes that something is always the focus of attention, and that evaluation of an attended item can take place concurrently with parallel processing of unattended items. The serial attention component of the search for the target occurs from the beginning of search, while all distractors are still in contention and undergoing parallel processing. CASPER provides a computational account of how serial processing can proceed concurrently with parallel processing during search, with the parallel process influencing the course of serial item selection by changing the priority according to which items are selected.

The second core theoretical claim is that parallel processing is necessarily more error-prone than serial processing of attended items. CASPER predicts high accuracy in the evaluation of a search item under attentional scrutiny, but the parallel evaluation process in CASPER is lower in accuracy.

A third core theoretical claim is that the distance from an item to fixation in the visual field has an impact on the visual processing of that item. This assumption manifests in multiple ways in the model. First, parallel processing has a stronger influence on selection prioritization the closer an item is to fixation. Items that are close to fixation undergo a greater degree of priority adjustment, while items in the far periphery tend to decay in priority. Item selection is also influenced by distance from fixation. The probability of an item being selected is governed by a version of Luce's (1959) choice axiom defined over both an item's selection priority and its distance from fixation. In this sense, attentional selection is neither deterministic nor purely random, which is the fourth core claim of the model.

The fifth core theoretical claim is that items can be accepted as the target only under attentional scrutiny. Parallel processing can change the likelihood of an item being selected for attentional scrutiny, and if parallel processing drives the selection priority of an item down below a rejection threshold, the parallel process will permanently remove the item from contention. However, the parallel process cannot determine that a search item is the target item. Items can be highly prioritized for selection, but selection and attentional scrutiny must occur before a final decision is made that the item is the target.

The representations in CASPER are simplified and preliminary, but the theoretical framework assumes that visual representations are hierarchical in nature (see Palmer 1977; Hummel & Biederman, 1992), and that while all levels of representation can contribute to the search process, the visual system will default to an entry level of representation if it is available to discriminate the target from the distractors (the sixth core theoretical claim). For example, some representations in CASPER are simple color and shape features (e.g., oriented lines, vertices, or hues) that might correspond to visual area V2, or perhaps V4. However, the representational scheme assumes that features that are relevant to finding objects (e.g., surfaces or parts) would dominate visual processing during search if the model were extended to support them (Enns & Rensink, 1991; Palmer & Rock, 1994)

However, this default assumption does not preclude the possibility that a level of representation other than parts or surfaces might dominate selection priority under some circumstances. What features are salient during search is a function of low-level salience, the features of the target and distractors, the instructions given to the participants, and any particular strategy a participant might adopt (the seventh claim). Although CASPER does not provide an explicit account of every one of these factors, the salience of a given level of representation can be adjusted in the model to compensate for the factors.

Finally, CASPER assumes representation and processing of relations can only be performed under attention (the eighth claim). This sets relations apart in the representational hierarchy. Representation of relations requires the binding of relational roles to their fillers on the fly (Hummel & Biederman, 1992), and Treisman & Gelade (1980) showed that binding requires attention. Therefore, CASPER does not predict a benefit of parallel processing for relations in visual search (see Logan, 1994). Instead, CASPER assumes that all features in a relational search are superimposed during parallel processing due to a lack of appropriate binding of relational roles to fillers. This superposition catastrophe impairs the discrimination of the target from distractors, necessitating a strict serial search.

| | |
|---|---|
| Claim #1 | *Serial and parallel processing occur concurrently and not in successive stages.* |
| Claim #2 | *Parallel processing is more error prone than serial processing.* |
| Claim #3 | *Visual processing near fixation has a greater impact on the evaluation of search items than processing in the periphery.* |
| Claim #4 | *Item selection is neither purely random nor deterministic, but is instead governed by the Luce choice axiom, where items most similar to the target and closest to fixation are more likely, but not guaranteed, to be selected for attentional scrutiny.* |
| Claim #5 | *Items are only accepted as the search target under attentional scrutiny, but they can sometimes be rejected as the target via parallel processing.* |
| Claim #6 | *Surfaces and parts act as entry level representations in search, absent other influences.* |
| Claim #7 | *Visual representations are hierarchical in nature and the level of representation that will dominate search depends on an interaction of top-down, bottom-up, and lateral influences, including but not limited to low-level visual salience and intentional search strategies on the part of participants.* |
| Claim #8 | *Processing of relations only happens with attention.* |

*Table 2.1 The core theoretical claims in the CASPER model of visual search*

**Overview of the Model**

CASPER uses concurrent serial and parallel processing to find the target in a visual search display, rather than the more typical two-stage sequence in which a parallel process finishes before the serial process begins. In CASPER, a serial process attends to a single search item at a time, comparing it to a *target template*. While this is happening, all the search items in the display update their priority for attentional selection in parallel.

At the beginning of search, all items in the display are set at a high priority for attentional selection. However, items immediately begin to decay in selection priority with each iteration of the model's simulation. At the same time, items are adjusted in selection priority in proportion to their similarity to the search target. Consistent with theoretical claim #2 that parallel processing is more error-prone, parallel priority updating includes a stochastic component. Specifically, rather than comparing all features of every search item to all the features in the search template during each iteration of parallel processing, CASPER compares only a probabilistically chosen selection of target features to the search items, as described later.

The contribution of parallel processing to the model's performance is not instantaneous; items are differentiated in priority via the parallel process over time as the model runs. Decay and boosting of selection priority occur continuously in parallel for all items in the display during search, except during eye movements. Items may become extremely unlikely to be selected for attentional scrutiny if they decay to a very low selection priority or be removed from contention completely if they fall below a minimum selection priority.

Items are accepted as the target via the serial process. When an item is selected for attentional scrutiny, the model compares the features of the target template to all the features of the selected search item. If the item is an exact match to the target template, then the target has been found and the search terminates. If the item is not an exact match to the target template, then the item is rejected (i.e., permanently removed from consideration as a search item) and a new item is selected for scrutiny until no candidate items remain in the display.

As the number of distractors in the display increases, there is more opportunity for items to decay out of contention via the parallel process before they are selected for attentional scrutiny. As a result of this process, CASPER produces negatively accelerating functions of response time (RT) vs. set size.

CASPER is tuned to simulate target-present search tasks. However, CASPER will detect a target-absent condition if all the items are marked as rejected via either the parallel or serial process. Tuning target-absent searches may require adjustments in search strategy on the part of participants, mechanisms for which are not currently incorporated into CASPER's account (see Chun & Wolfe, 1996 for a discussion of the complexities of target-absent searches). Accordingly, only simulations of target present searches are reported in this project.

**Representation of Shape and Color in CASPER**

CASPER encodes both the target template and the search items as vectors of features that represent aspects of color and shape. As noted in theoretical claim #6, CASPER assumes that visual properties represented in neurons as early as V2 can serve as features for search, but neurons in earlier visual areas (e.g., V1 or LGN) cannot. The model further assumes that the visual system will utilize all available levels of representation that are appropriate to the task (e.g., line segments and junctions in V2, color and surfaces in V4, and object parts and whole objects in lateral occipital complex (LOC) and higher areas), but that higher-order representations are by default the most highly prioritized. In the most basic version of the model, all the representations of shape are a hybrid of low and mid-level features that are assumed to contribute to the representation of shape. However, no claims are made about their completeness or correctness. They are better viewed as a first approximation and a work in progress that is extensible to more sophisticated representational schemes in the future.

The simplest representations of shape in the current model include oriented line segments, L-vertices, T-junctions, and X-junctions. These can be combined to create typical visual search stimuli, such as Ts, Ls, Xs, Os, and so forth. In later simulations in Chapter 4, more sophisticated representations are added as might be found in V4 or LOC. CASPER represents color using an opponent color system like that of the primate visual system. Color is coded according to red-green opponency, blue-yellow opponency, and black-white opponency.

Each feature is represented as a trinary (1, 0, -1) encoding of presence (1), absence (0), or opposition (-1) in either the target template or search items. Some features are coded by multiple vector entries, allowing CASPER to encode the strength of a feature's presence (e.g., a saturated red vs. an unsaturated red). Extra representational resolution is given to vertically and horizontally oriented line segments. The representations of color and shape in the basic version of the model are shown in Tables 2.2 and 2.3, respectively.

By default, all features, $k$, have the same level of salience, $\eta_k = 1.0$. However, there are some cases in which a feature's salience might need to be different than the default.[5] A multiplicative salience parameter, $\eta_k$, can tune the salience of any feature, $k$, to better fit the empirical data. Changing the value of $\eta_k$ from the default value of 1.0 corresponds to the assumption that, in the simulation in question, feature $k$ is either more or less salient than it would tend to be otherwise. For example, $\eta_k$ might deviate from the default in an experiment in which the subject was explicitly instructed to attend to feature $k$. CASPER provides no account of top-down cognitive biases on search, so the salience parameter can be used to compensate for such effects when they are hypothesized to be a factor. Another reason that salience might be adjusted is that CASPER has simplified representations that cannot capture the full range of shape and color properties. A simulation that is dependent on subtle differences in color may need fine adjustments because only 18 units are responsible for representing color.

---

[5] Thank you to Diane Beck for her comments that led to the insight that salience should be an explicit parameter that can be used to bias the interaction of top-down and bottom-up influences during search.

Changing the salience parameter in CASPER changes the degree of contribution to the similarity calculation during parallel processing. This mechanism can have a similar net effect to the redundant coding of features in CASPER (i.e., as is the case with color and orientation units). Even though representational strength and salience have similar effects in CASPER, I do not take the theoretical position that they are necessarily the same thing in the human visual system, although I consider this is to be a possibility.

| Color | White/Black | Red/Green | Blue/Yellow |
|---|---|---|---|
| White | 1, 1, 1,-1,-1,-1 | 0, 0, 0, 0, 0, 0 | 0, 0, 0, 0, 0, 0 |
| Black | -1,-1,-1, 1, 1, 1 | 0, 0, 0, 0, 0, 0 | 0, 0, 0 ,0, 0, 0 |
| Red | 0, 0, 0, 0, 0, 0 | 1, 1, 1,-1,-1,-1 | 0, 0, 0 ,0, 0, 0 |
| Green | 0, 0, 0, 0, 0, 0 | -1,-1,-1, 1, 1, 1 | 0, 0, 0 ,0, 0, 0 |
| Blue | 0, 0, 0, 0, 0, 0 | 0, 0, 0, 0, 0, 0 | 1, 1, 1,-1,-1,-1 |
| Light Blue | 1, 1, 1, 1, 1, 1 | 0, 0, 0, 0, 0, 0 | 1, 1, 1,-1,-1,-1 |
| Yellow | 0, 0, 0, 0, 0, 0 | 1, 1, 1, 1, 1, 1 | 0, 0, 0, 1, 1, 1 |
| Orange | 0, 0, 0, 0, 0, 0 | 1, 1, 0,-1,-1, 0 | -1, 0, 0, 1, 0, 0 |
| Pink | 1, 1, 0,-1,-1, 0 | 1, 0, 0,-1, 0, 0 | 0, 0, 0 ,0, 0, 0 |
| Dark Green | 1, 1, 1,-1,-1, 0 | -1,-1,-1, 1, 1, 1 | 0, 0, 0, 0, 0, 0 |
| Brown | 1, 1, 1,-1,-1, 0 | 1, 1,-1,-1,-1, 1 | 0, 0, 0, 0, 0, 0 |

*Table 2.2 Representations of color in CASPER. Color is represented using an opponent encoding scheme with white/black, red/green, and blue/yellow channels. Each of the six basic color categories used to encode opponent color have three units of information and can take on three values: 1, 0, and -1. For example, a strongly red value would be encoded as {1, 1, 1,-1,-1,-1} on the red/green channel to indicate maximum red activity and the absence of green activity. An orange stimulus has a red/green channel encoding of {1, 1, 0, -1, -1, 0} to indicate the presence of some red information without the presence of green, and a blue/yellow encoding of {-1, 0, 0, 1, 0, 0} to indicate the presence of some yellow without the presence of blue.*

| Shape | Orientation coding in units of π/8 radians | | | | | | | | L-vertex | T-junction | X-junction |
|---|---|---|---|---|---|---|---|---|---|---|---|
| | 0 | 1 | 2 | 3 | 4 | 5 | 6 | 7 | | | |
| Horizontal | 1, 1, 1 | 1, 0 | 0, 0 | 0, 0 | 0, 0, 0 | 0, 0 | 0, 0 | 1, 0 | 0, 0, 0, 0 | 0, 0, 0, 0 | 0 |
| Vertical | 0, 0, 0 | 0, 0 | 0, 0 | 1, 0 | 1, 1, 1 | 1, 0 | 0, 0 | 0, 0 | 0, 0, 0, 0 | 0, 0, 0, 0 | 0 |
| π/4 Diagonal | 0, 0, 0 | 1, 0 | 1, 1 | 1, 0 | 0, 0, 0 | 0, 0 | 0, 0 | 0, 0 | 0, 0, 0, 0 | 0, 0, 0, 0 | 0 |
| 3π/4 Diagonal | 0, 0, 0 | 0, 0 | 0, 0 | 0, 0 | 0, 0, 0 | 1, 0 | 1, 1 | 1, 0 | 0, 0, 0, 0 | 0, 0, 0, 0 | 0 |
| T1 | 1, 1, 1 | 1, 0 | 0, 0 | 1, 0 | 1, 1, 1 | 1, 0 | 0, 0 | 1, 0 | 0, 0, 0, 0 | 1, 0, 0, 0 | 0 |
| T2 | 1, 1, 1 | 1, 0 | 0, 0 | 1, 0 | 1, 1, 1 | 1, 0 | 0, 0 | 1, 0 | 0, 0, 0, 0 | 0, 1, 0, 0 | 0 |
| T3 | 1, 1, 1 | 1, 0 | 0, 0 | 1, 0 | 1, 1, 1 | 1, 0 | 0, 0 | 1, 0 | 0, 0, 0, 0 | 0, 0, 1, 0 | 0 |
| T4 | 1, 1, 1 | 1, 0 | 0, 0 | 1, 0 | 1, 1, 1 | 1, 0 | 0, 0 | 1, 0 | 0, 0, 0, 0 | 0, 0, 0, 1 | 0 |
| L1 | 1, 1, 1 | 1, 0 | 0, 0 | 1, 0 | 1, 1, 1 | 0, 0 | 0, 0 | 1, 0 | 1, 0, 0, 0 | 0, 0, 0, 0 | 0 |
| L2 | 1, 1, 1 | 1, 0 | 0, 0 | 1, 0 | 1, 1, 1 | 0, 0 | 0, 0 | 1, 0 | 0, 1, 0, 0 | 0, 0, 0, 0 | 0 |
| L3 | 1, 1, 1 | 1, 0 | 0, 0 | 1, 0 | 1, 1, 1 | 0, 0 | 0, 0 | 1, 0 | 0, 0, 1, 0 | 0, 0, 0, 0 | 0 |
| L4 | 1, 1, 1 | 1, 0 | 0, 0 | 1, 0 | 1, 1, 1 | 0, 0 | 0, 0 | 1, 0 | 0, 0, 0, 1 | 0, 0, 0, 0 | 0 |
| X | 0, 0, 0 | 1, 0 | 1, 1 | 1, 0 | 0, 0, 0 | 1, 0 | 1, 1 | 1, 0 | 0, 0, 0, 0 | 0, 0, 0, 0 | 1 |
| O | 1, 0, 0 | 1, 0 | 0, 0 | 1, 0 | 1, 0, 0 | 1, 0 | 1, 0 | 0, 0 | 0, 0, 0, 0 | 0, 0, 0, 0 | 0 |
| Q | 1, 0, 0 | 1, 0 | 0, 0 | 1, 0 | 1, 1, 1 | 1, 0 | 1, 0 | 0, 0 | 0, 0, 0, 0 | 0, 0, 0, 0 | 0 |
| G1 | 1, 1, 1 | 1, 1 | 1, 1 | 1, 0 | 1, 1, 1 | 1, 0 | 0, 0 | 1, 0 | 1, 0, 0, 0 | 0, 0, 0, 0 | 1 |
| G2 | 1, 1, 1 | 1, 0 | 0, 0 | 1, 0 | 1, 1, 1 | 1, 1 | 1, 1 | 1, 0 | 1, 0, 0, 0 | 0, 0, 0, 0 | 0 |
| P1 | 1, 1, 1 | 1, 1 | 1, 1 | 1, 1 | 1, 1, 1 | 1, 0 | 0, 0 | 1, 0 | 1, 0, 1, 0 | 0, 0, 0, 0 | 1 |
| P2 | 1, 1, 1 | 1, 0 | 0, 0 | 1, 0 | 1, 1, 1 | 1, 1 | 1, 1 | 1, 0 | 1, 0, 1, 0 | 0, 0, 0, 0 | 0 |

*Table 2.3 The simplified representations of low-level shape in CASPER. In the current version of the model, shape is represented using four basic elements of shapes: oriented line segments in increments of π/8 radians, L-vertices, T-junctions, and X-junctions. Orientation is coded with multiple units to capture the strength of the representation, where all units on corresponds to a strong representation and all units off is the absence of a feature. Horizontal and vertical segments have an extra unit to capture bias toward horizontal and vertical orientations. Oriented segments can have overlap in orientation. For example, a diagonal corresponding to orientation π/4 might also have weak activity in π/8 and 3π/8 orientations as well. L-vertices and T-junctions units correspond to four possible orientations. X junctions have only one orientation. The model shapes include oriented line segments, Ts in 4 orientations, Ls in 4 orientations, X, O, Q-like shapes, and stimuli that correspond to Pomerantz et al. (1977). Shapes G1 and G2 are the low-level encodings of shapes from the Good configural context condition from Pomerantz et al. (1977) and shapes P1 and P2 are the low-level encodings of shapes from the Poor configural context condition.*

**Item Selection in CASPER**

Unlike some other models of visual search, (e.g., Target Contrast Signal Theory; Lleras et al., 2020), candidate search items in CASPER are not selected strictly at random for attentional scrutiny. Nor is the most active item deterministically selected for attentional scrutiny (e.g., as in GS; Wolfe et al., 1989). Instead, similar to Bundesen's (1990; 1998) Theory of Visual Attention (TVA), search items are selected for serial scrutiny in CASPER according to Luce's (1959) choice axiom, where the probability $P(i)$ of an item $i$ being selected from the set of remaining items $I$ is given by:

$$P(i) = \frac{p_i * \varepsilon_i}{\sum_i p_i * \varepsilon_i} \quad (1)$$

where $p_i$ is the selection priority of a given item as determined by the parallel processing mechanism described below, and $\varepsilon_i$ is a weight that is based on the distance of an item from fixation. This means that the item with the highest selection priority $p_i$ that is close to fixation is likely, but not guaranteed, to be the next item selected for attentional scrutiny. The distance weight, $\varepsilon_i$, is given by

$$\varepsilon_i = 1 - D_i/D_{max} \quad (2)$$

where $D_i$ is the distance from fixation to $i$, and $D_{max}$ is a parameter that determines the distance at which $\varepsilon_i$ goes to zero (i.e., the item is considered too distant to undergo visual processing). By default, $D_{max}$ = 4.0 visual field radii. [6]

The model can be permitted to move its eyes (i.e., changing the location of fixation) to the newly selected item, $i$, by setting a parameter. By default, eye movements are permitted. If the model is allowed to move its eyes, then the likelihood that it will fixate the selected item is equal to $\varepsilon_i$.

Eye movements incur a time cost of 30 simulation iterations, during which time parallel processing does not proceed.[6] The cost of an eye movement is 15 times the cost of a covert attentional shift, which is only two simulation iterations. Item selection priorities are retained across eye movements (see Lleras et al., 2005). Once an item has been selected, attentional scrutiny occurs concurrently with parallel processing, meaning that additional parallel processing does not wait for serial processing to finish to proceed.

As items are rejected via serial scrutiny, the set of items $I$ becomes progressively smaller until either the target is found, or all items have been selected and rejected.

---

[6] Parameter values were chosen because they produced the best fits to the human data during simulations and are plausible in terms of the number of equivalent milliseconds per simulation iteration (reported in Chapter 3) and other factors. The parameter values are constant across all simulations reported in Chapters 3 and 4.

**Parallel Processing in CASPER**

Parallel processing occurs continuously during search, and its primary function is to set the priorities of the search items for attentional scrutiny. Search items' priorities start at a value of $p_i$=1.0 + $\omega$, where $\omega$ is a random number drawn from a uniform distribution between -0.1 and 0.1. Priorities are updated individually on every iteration of a simulation run (except during eye movements). The change, $\Delta p_i$, of item $i$'s priority for selection is based on two components. The first is a decay factor, $\delta$, that is uniformly applied over all items on every iteration of a simulation run (except during eye movements). The second component is an adjustment that either boosts or penalizes $p_i$ as a function of $i$'s similarity to the target template. On each iteration (except during eye movements), the change in the selection priority of any item $i$ is given by:

$$\Delta p_i = p_i * (1 - \delta) + \varepsilon_i * \phi(i, T) * \rho \qquad (3)$$

where $\delta$ = 0.5, $\varepsilon_i$ is the effect of distance from fixation given in Equation (2), $\phi(i, T)$ is the similarity of the item $i$ to the template $T$ (see Equation (4) below), and $\rho$ is a random number taken from a uniform distribution between 0 and 1 that approximates noise in the system. When $p_i$ falls below a minimum selection priority, $p_{min}$ = 0.001, the item is considered rejected. [6]

Consistent with theoretical claim #3, the model assumes that the closer an item is to fixation, the greater the ability to compare that item to the target template during parallel processing. This is instantiated by multiplying the match calculation, $\phi(i,T) * \rho$, in the second term of Equation (3) by the distance weight, $\varepsilon_i$.

The similarity match in the model is governed by a probabilistic sampling of feature dimensions $k$ in both the target template and the search items. At the beginning of a simulation, all possible feature dimensions are categorized as either *relevant*, *irrelevant*, or *absent*. A feature is considered relevant if there is a mismatch on that feature between the target template and any search item (i.e., this feature is relevant to the task of discriminating the target from the distractors). A feature is deemed irrelevant if it is present in both the target and all the non-targets (i.e., it is present but not relevant to the problem of discriminating the target from the distractors). A feature is deemed absent if no search item contains it. Relevant features are sampled with greater probability (probability = 0.85) than features that are irrelevant (probability = 0.15).[6] Features that are absent (i.e., have a value of zero) are not sampled.

The feature match during parallel processing is the sum of the match quality across all randomly chosen sampled features divided by the total number of relevant features, given by:

$$\phi(i,T) = \frac{\tau * \sum_k s(k) m(i_k, T_k) w_k \eta_k}{\sum_k r_k} \tag{4}$$

where $\tau$ is a constant multiplier that governs the strength of parallel processing (equal to 3.0) [6,7], $s(k) = 1$ if feature $k$ was sampled and 0 if it was not, and $r_k$ is equal to the salience, $\eta_k$, of feature $k$ if $k$ is relevant and 0 if it is not. The feature weight in the match calculation, $w_k = 3.0$ if the feature is present in the search template. [6] If a feature is absent in the target template but present in the search item, $w_k = 0.1$. [6] The match, $m(i_k, T_k)$, is +1 if the value of $k$ in the search template matches the value of $k$ in item $i$, and -1 if they mismatch. Recall that in the current version of the model, all features are trinary, with a value of 1 if they are present, 0 if they are absent, and -1 if they are opposed by a present feature (e.g., in the way that green is opposed by red). Any mismatch in sign or magnitude counts the same toward the match calculation.

In other words, if a feature is sampled and there is a match between the target template and the search item, positive evidence is accumulated. If there is not a match, negative evidence is accumulated. If the sampled feature is present in the target template, it will count more toward the match or mismatch between the target template

---

[7] Thank you to Benjamin Evans for his comments about the possible return of rejected items that ultimately led to this change in the algorithm. Although the model does not ultimately incorporate a mechanism by which rejected items can return, considering the implications of such a mechanism led to the insight that there needs to be a greater impact of parallel processing between attentional selections in order to properly predict difficult searches.

and the selected item than if the feature is in the search item but not in the search template.

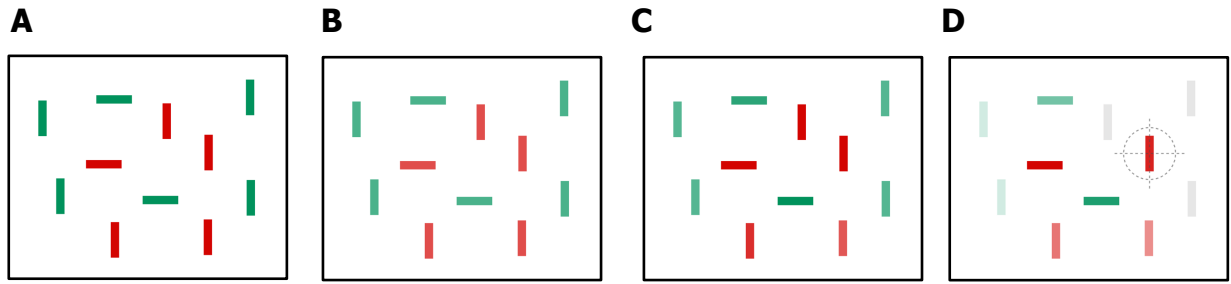

*Figure 2.1 A hypothetical search display in which the target is a horizontal red line. Items with higher priority are depicted in more saturated colors. A) At the beginning of search, all items are high priority for selection for focused attentional scrutiny. (B) Selection priority immediately begins to decay for all items. (C) Item priority is stochastically adjusted in proportion to each item's similarity to the search target and its distance from fixation. (D) Decay and adjustment occur continuously in parallel across all items during search.*

Figure 2.1 illustrates the parallel search process. At the beginning of a search (Figure 2.1A) all item priorities start high and immediately begin to decay (Figure 2.1B). Items that are near fixation will undergo greater adjustment than items that are far from fixation. Over time, items that are a better match to the target template are adjusted to higher priority compared to items that are worse matches (Figure 2.1C). Some items that are not good matches and are far from fixation will rapidly lose selection priority so that they are rejected and removed from contention permanently (Figure 2.1D)

**Serial Processing in CASPER**

The serial process of selecting items for attentional scrutiny is depicted in Figure 2.2. CASPER focuses attentional scrutiny on one item at a time (Figure 2.2A). The match between the target template and a search item does not have a probabilistic component like in the parallel process. Instead, the match under attentional scrutiny is computed as a strict match on all the features between the target template, $T$, and the attended item, $i$. If an item is not an exact match for the target template, then the item is rejected and permanently removed as a candidate (Figure 2.2B). It will no longer be selected by the model for attentional scrutiny, nor will it be updated via the parallel process (Figure 2.2C). If the target is found, the search terminates (Figure 2.2D). If all the items are rejected, the search also terminates.

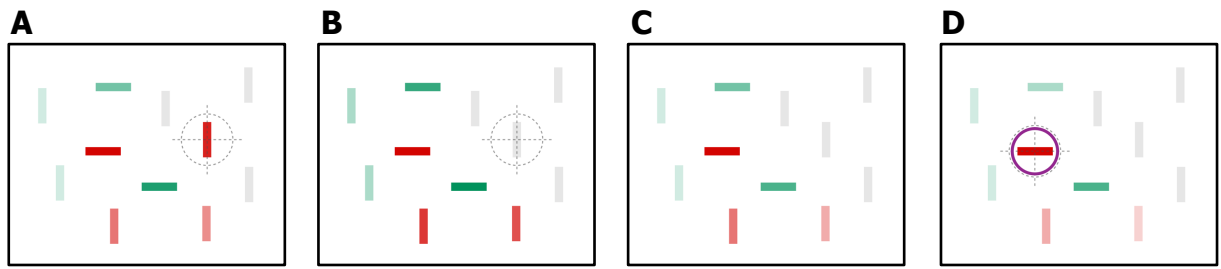

*Figure 2.2 Serial processing in the same search for a red horizontal line. A) While the parallel process continues, items are serially selected for focused attention (selection is depicted by the gray dotted circle) via Luce's choice axiom. B) Selected items are matched to the target or rejected by strict comparison across all feature dimensions. C) Rejected items are removed from contention for future selection. D) When the search target is selected and matched via focused attention (indicated by the purple circle), or all the items have been scrutinized, the search terminates.*

# Chapter 3: Simulations

## Simulation 1: Feature vs. conjunction search

Treisman & Gelade (1980) Experiment 1 showed that when a search target can be discriminated from distractors on the basis of a single feature, search can proceed rapidly and in parallel. In the Color Feature condition of their experiment, a search for a blue X among brown Ts and green Xs produced a shallow search slope (Figure 3.1A). However, when the target item shared features of both distractors (i.e., a green T among brown Ts and green Xs) search was difficult, producing a nonzero slope of 28.7ms/item.

*Methods*

For Simulation 1, CASPER was given two conditions corresponding to the feature and conjunction search conditions of Treisman & Gelade (1980) Experiment 1. In the Feature condition, the search target was a blue X among dark green Xs and brown Ts. In the Conjunction condition, the search target was a green T among dark green Xs and brown Ts, as defined according to the representations in Tables 2.2 and 2.3. The model was given set sizes 1 (target-only), 5, 15, and 30, which correspond to the original experiment. Each simulation consisted of 52 trials per set size per condition. The model was run 100 times (i.e., for 100 full simulations), corresponding to 100 virtual subjects. Response times in units of model iterations were recorded.

*Results and discussion*

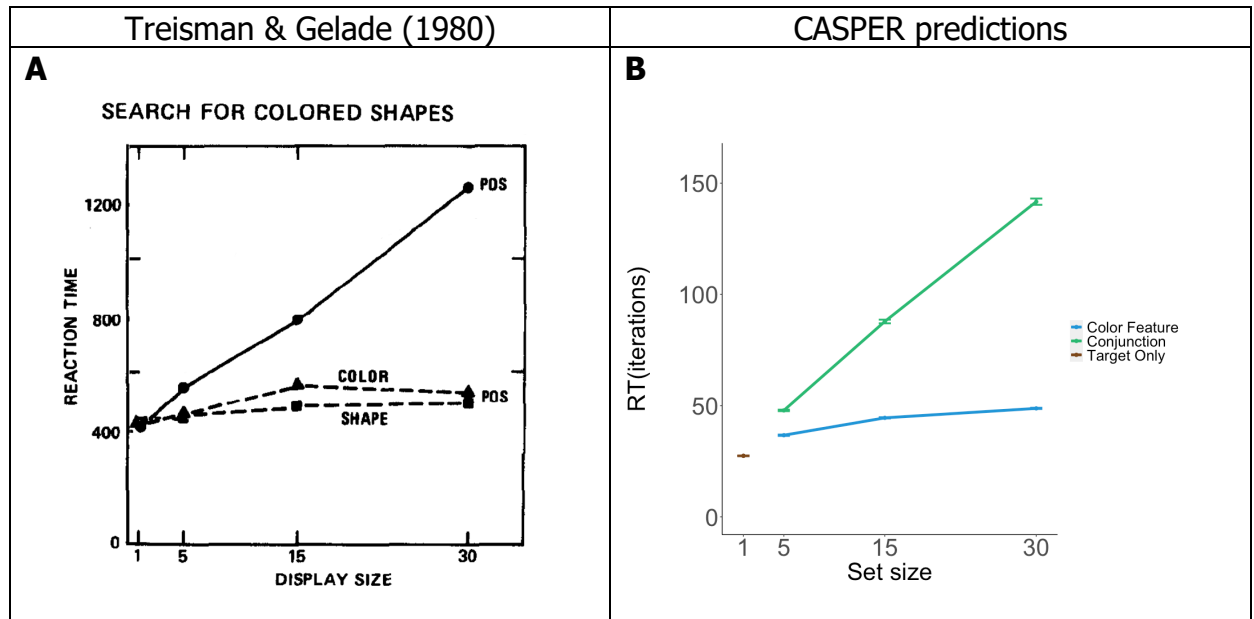

*Figure 3.1 A) Figure modified from Treisman & Gelade (1980) to show only target present conditions. The color feature search produces a function of response time to set size that features a flat slope, meaning that there is not substantial additional processing time required as display set size increases. Conjunction search produces a function of response time vs. set size that features a steep positive slope of 28.7 additional milliseconds per search item for display processing (corresponding to 57.4ms of processing time for each item because on average the target is found halfway through the search). B) The CASPER model produces the same pattern of results obtained by Treisman & Gelade for feature and conjunction search, Error bars are the standard error of the mean. Reprinted from Cognitive psychology, 12(1), 97-136 Treisman, A. M., & Gelade, G. (1980). A feature-integration theory of attention. Copyright 1980, with permission from Elsevier*

The mean response time for each virtual subject per condition and set size was computed, and the mean over virtual subjects of the response time per set size per condition is depicted in Figure 3.1B. The model produced a qualitative match to the pattern of data obtained in the original experiment, with a nearly flat search slope for the feature condition, and a steep search slope for the conjunction condition. The model underpredicted the difficulty of the target-only condition, responding faster (relative to display sizes five and above) than the human subjects.

Coefficients of determination were computed to assess the model fit to set sizes other than the target-only condition. The coefficient of determination was $R^2 = 0.922$ for the feature condition and $R^2 = 0.999$ for the conjunction condition. The combined fit, obtained by concatenating the simulation data from the two conditions into a single vector and comparing it with similarly concatenated human data, was $R^2 = 0.998$.

The estimated number of milliseconds per simulation iteration can be computed for each condition by dividing the range of the human data within each condition by the range of the model data within each condition, excluding the target-only condition. Each iteration of the CASPER simulations corresponds to an estimated 4.5ms and 6.3ms in the human data for the Feature and Conjunction conditions, respectively, indicating that the model either slightly underestimates conjunction search difficulty or overestimates feature search difficulty: the greater the milliseconds per iteration, the longer the human subject takes per iteration in CASPER, meaning that CASPER is more efficient compared to the human. Ideally, the estimates of milliseconds per iteration would be the same for both conditions.

As described in Chapter 2, CASPER simulates concurrent serial and parallel processing during visual search. In the Feature condition, the blue target has few color or shape features in common with the distractors, making differentiation during parallel processing more efficient. Distractors undergo decay and are also simultaneously adjusted downward in selection priority due to their poor match to the target template. Meanwhile, the target is boosted in selection priority due to its good match to features in the target template.

In the Conjunction condition, the distractors have greater feature overlap with the target template, meaning that they do not decay in selection priority as rapidly as they do in the Feature condition, leading to a greater likelihood that a nontarget item will be selected for attentional scrutiny instead of the target.

CASPER underestimates the difficulty of the target-only condition for each type of search. The original experiment included both target-present and target-absent trials. The discrepancy between the model and the empirical data for the target-only condition may be due to a response selection effect that CASPER does not account for.

**Simulation 2: Conjunction search with nonoverlapping distractor features**

Wolfe et al. (1989) showed that subjects can perform conjunction searches more efficiently than was predicted by FIT. In their Experiment 2, subjects searched for a green horizontal target among red horizontal and green vertical distractors. In this experiment, subjects were divided into three groups, with each group performing a search over different numbers of set sizes. In each group, Wolfe et al. demonstrated a nonlinearity in the function of response time vs. set size, whereby searches over larger set sizes appeared to be more efficient than searches over smaller set sizes (Figure 3.2A).

*Methods*

For Simulation 2, CASPER was given two conditions. The first condition corresponded to the Conjunction search conditions in Wolfe et al. (1989) Experiment 2. A second Feature condition that did not exist in the original experiment was added as a point of reference. In the Conjunction condition, the search target was a green horizontal bar among green vertical bars and red horizontal bars, for which the representational encodings can be found in Tables 2.2 and 2.3. In the Feature condition, the search target was a green horizontal bar among red vertical bars. The model was given set sizes designed to span the range of set sizes used by Wolfe et al.: 1 (target-only), 3, 5, 9, 17, 25, and 37. Each simulation consisted of 52 trials per set size per condition. The model was run 100 times (i.e., for 100 full simulations), corresponding to 100 virtual subjects. Response times in units of model iterations were recorded.

*Results and Discussion*

The number of subjects per group in the original experiment was small, and thus estimates of the actual numerical values of the mean response time per set size per condition are imprecise, making it difficult to calculate a correlation between CASPER's response times and those of the human participants. For this reason, in Simulation 2, CASPER is only used to show a qualitative match to the experiment data.

The mean response time for each virtual subject per condition was computed, and the mean over virtual subjects per set size per condition is depicted in Figure 3.2B. Simulation of both the Feature and Conjunction conditions produces a negatively accelerating function of reaction time vs. set size, but a simple feature search is still more efficient than the conjunction search with nonoverlapping features.

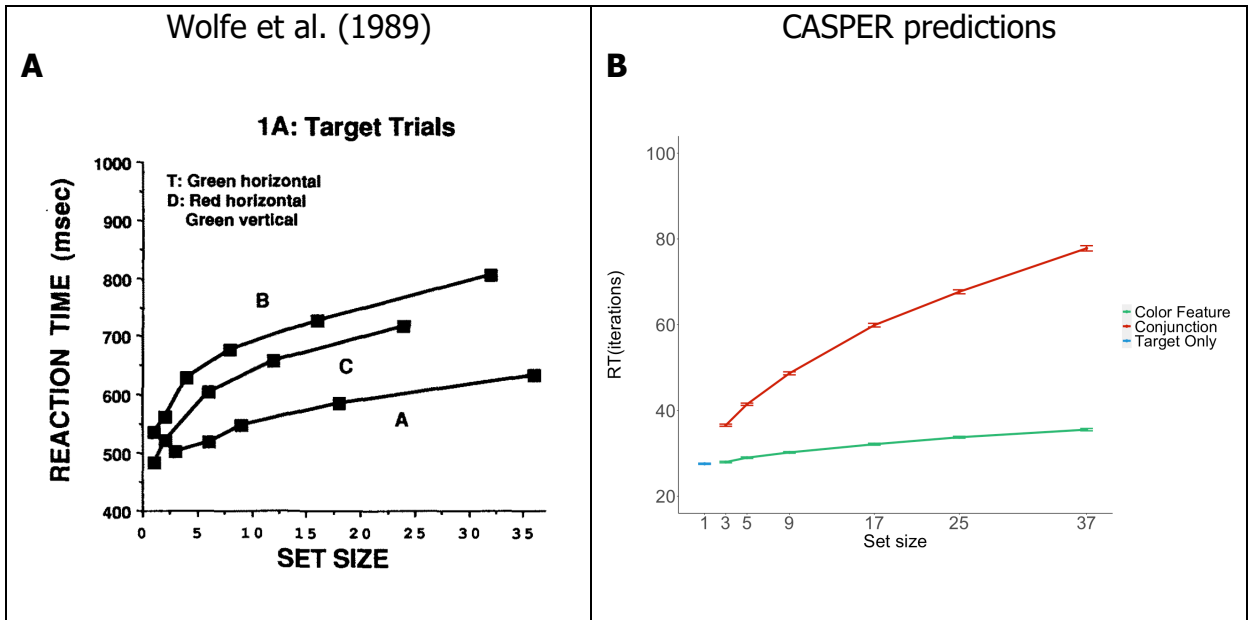

Figure 3.2 A) Figure from Wolfe et al. (1989) Experiment 2, showing conjunction search results for three groups of subjects with three different groups of set sizes.  B) The CASPER model produces the same pattern of results obtained by Wolfe et al.'s conjunction searches in their Experiment 2.  Error bars are the standard error of the mean. Copyright © 1989 by the American Psychological Association. Reproduced with permission. Wolfe, J. M., Cave, K. R., & Franzel, S. L. (1989). Guided search: An alternative to the feature integration model for visual search. Journal of Experimental Psychology: Human Perception and Performance, 15(3), 419–433.

Unlike the result of the simulation of the conjunction search in Treisman & Gelade (1980) Experiment 1, CASPER produces a negatively accelerating function of response time vs set size for the conjunction search in Wolfe et al. (1989) Experiment 2. CASPER's improved search efficiency in this case can be explained by the representations of the experimental stimuli. The more feature overlap between the target template and the distractors, the more difficult a search is for CASPER.

|  | Treisman & Gelade (1980) Experiment 1 | Wolfe et al. (1989) Experiment 2 |
|---|---|---|
| Distractor 1 features in common with the target template | *11 color*<br>*4 shape* | *6 color*<br>*0 shape* |
| Distractor 1 features that differed from the target template | *0 color*<br>*11 shape* | *10 shape*<br>*0 color* |
| Distractor 2 features in common with the target template | *7 color*<br>*11 shape* | *0 color*<br>*5 shape* |
| Distractor 2 features that differed from the target template | *4 color*<br>*0 shape* | *6 color*<br>*0 shape* |

*Table 3.1 A comparison of the degree of feature overlap between the target template and the distractors in Treisman & Gelade (1980) Experiment 1 and Wolfe et al. (1989) Experiment 2, as encoded in CASPER. For the Treisman & Gelade simulation, CASPER finds more features in common between the target and distractors than in the Wolfe et al. simulation, which results in a less efficient separation of selection priority between the target and distractor items.*

In the case of Wolfe et al.'s (1989) experiment, the features in the distractors have less in common with the search target than the distractors used in Treisman & Gelade's Experiment 1. In CASPER's representation of Treisman & Gelade's stimuli, brown and dark green color representations have feature overlap, as do X and T shape representations. By comparison, the representation of red in CASPER has no feature overlap with the representation of green and the representation of vertical has no overlap with the representation of horizontal. The degree of feature overlap in CASPER's representations for the stimuli for Treisman & Gelade (1980) and Wolfe et al. (1989) are shown in Table 3.1.

**Simulation 3: Target-distractor similarity for feature search**

Target-distractor similarity not only affects conjunction search, but also has an impact on feature search. Like Wolfe et al. (1989), Buetti et al. (2016) demonstrated negatively accelerating functions of RT vs set size when search was performed over a large range of set sizes, but in feature search. Buetti and colleagues further showed that varying target-distractor similarity over both color and shape results in similarly varying efficiency in feature search, with some searches proceeding with a large impact of parallel processing, while other searches proceed less efficiently, with a greater log slope and less negative acceleration as set size increases (Figure 3.3A).

*Methods*

In Experiment 1A, Buetti et al. used a single target, a red triangle, across conditions designed to vary the similarity of the distractors to the target along both color and shape. In a low-similarity feature search condition, the distractors were blue circles. In a second high-similarity condition, the distractors were orange diamonds. In a third intermediate condition, the distractors were yellow triangles.

CASPER's current representations do not include shapes like diamonds and triangles. Therefore, for the simulations of Buetti et al. (2016) Experiment 1A, simplified representations of shape were used. In each simulation, the target was a red vertical line. In the low similarity condition, the distractors were light blue horizontal lines, which have no feature overlap with red verticals. In the high-similarity condition, the distractors were orange verticals, which have a high degree of overlap with red verticals. In the intermediate condition, the distractors were yellow verticals, which have some feature overlap with red verticals, but not as much overlap in color feature dimensions as orange verticals. To compensate for the smaller number of features in the simplified representations of shape in the simulations compared to the stimuli in the experiment, the salience of shape information was increased to 1.5 times the salience of shape.[8]

The model was given set sizes 1 (target-only), 2, 5, 10, 20, and 32, which correspond to the set sizes in the original experiment. Each simulation consisted of 52 trials per set size per condition. The model was run 100 times (i.e., for 100 full simulations), corresponding to 100 virtual subjects. Response times in units of model iterations were recorded.

*Results and Discussion*

The mean over the virtual subjects' mean response time per set size per condition is depicted in Figure 3.3B. CASPER shows a good qualitative match to the results of the Buetti et al. (2016) experiment, with the low-similarity search condition proceeding most efficiently, the high-similarity condition proceeding least efficiently, and the intermediate-similarity condition falling in the middle. All conditions produce negatively accelerating functions of RT vs set size.

---

[8] Making shape more salient provides a better quantitative fit to the empirical data but does not otherwise change the qualitative predictions of the model.

A coefficient of determination, $R^2$, was calculated between the empirical data and the simulation results for each condition to assess fit. For the high-similarity condition, $R^2$ = 0.994. For the intermediate-similarity condition, $R^2$ = 0.955. For the low-similarity condition, $R^2$ = 0.852. The combined total fit for the experiment as a whole, obtained by concatenating the simulation data from the three conditions into a single vector and comparing it with similarly concatenated human data, was $R^2$ = 0.926.

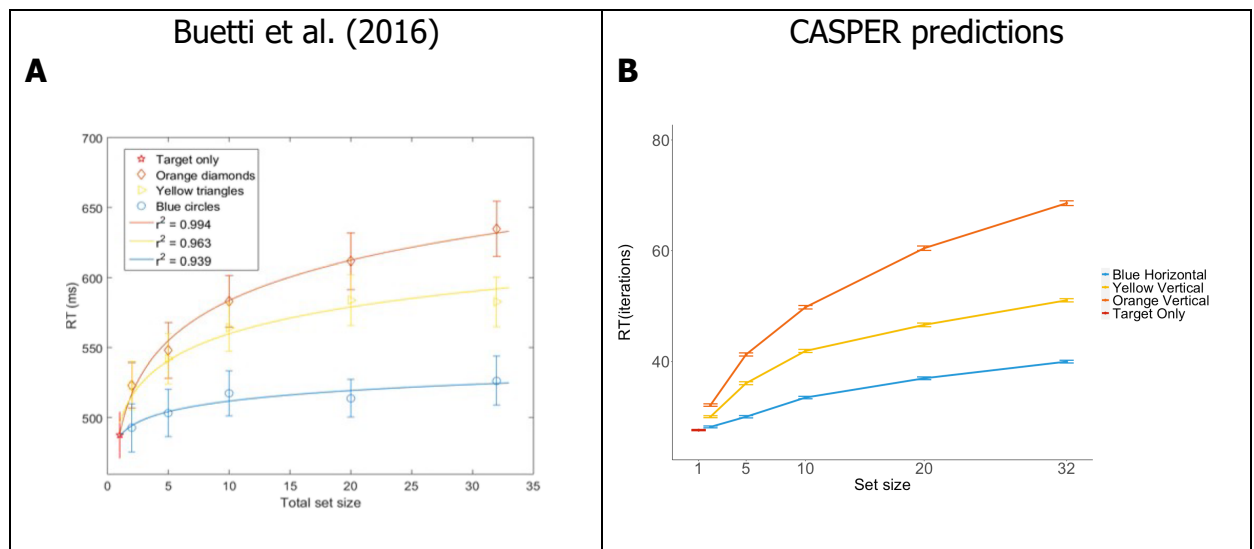

*Figure 3.3 A) Buetti et al. (2016) Experiment 1A, showing response times as a function of set size for high-, intermediate-, and low-similarity distractors that varied in both shape and color from a single target (a red triangle) across all conditions B) CASPER's predictions using stimuli designed to mimic the Buetti et al. stimuli. The model produces a qualitative match to the empirical results. Figure 3.3 A Copyright © 2020 Springer Nature. Reproduced with permission. Lleras, A., Wang, Z., Ng, G. J. P., Ballew, K., Xu, J., & Buetti, S. (2020). A target contrast signal theory of parallel processing in goal-directed search. Attention, Perception, & Psychophysics, 82, 394-425.*

The estimated number of milliseconds per simulation iteration was computed for each condition by dividing the range of the human data within each condition by the range of the model data within each condition, excluding the target-only condition. Each simulation iteration in CASPER corresponds to 2.72ms, 2.61ms, and 2.69ms for the high-, intermediate-, and low-similarity conditions, respectively.

CASPER produces better quantitative fits to the high- and intermediate-similarity conditions than to the low-similarity condition. The stimuli used in this simulation were simplifications of the Buetti et al. (2016) stimuli, so it's possible that the relative differences in target-distractor similarity were not captured by the model's representations, and that higher-order shape representations like diamonds and triangles could differentiate easy from difficult feature search. Another possibility is that CASPER's representation of color doesn't capture the difference in color space between the red target and the light blue distractor in the low-similarity condition. A third possibility is that CASPER's parameters and similarity match algorithm need further tuning to better characterize easy searches.

**Simulation 4: Search asymmetries**

Treisman & Souther (1985) Experiment 1 showed that reversing a target and distractors can change the difficulty of search when one type of search item has additional features beyond those of the other type of search item. For example, a search for a Q among Os proceeds more efficiently than a search for an O among Qs, presumably because Qs have an extra line segment in addition to the features of the Os (Figure 3.4A). When the search target has only a subset of the features of the distractors, search is difficult. When the search target has a superset of the distractor features, search is efficient. Treisman & Souther interpreted such asymmetries in search difficulty to mean that the features of the target template that are relevant to search are processed preattentively, whereas features that are not present in the target template are not able to be processed preattentively.

*Methods*

For Simulation 4, CASPER was given two conditions corresponding to the conditions of Treisman & Souther (1985) Experiment 1. In the Extra Feature condition, the search target was a white Q-like shape that featured a vertical line intersecting the O at the bottom (rather than a diagonal line as in a typical letter Q; For simplicity, this will be referred to simply as 'Q' going forward). The distractors were white Os. In the No Extra Feature condition, the search target was a white O among white Qs. The representation of the Q in CASPER includes two vertically oriented lines as additional features that are not present in the representation of an O.

The model was given set sizes 1 (target-only), 6, and 12, which correspond to the original experiment. Each simulation consisted of 52 trials per set size per condition. The model was run 100 times (i.e., for 100 full simulations), corresponding to 100 virtual subjects. Response times in units of model iterations were recorded.

*Results and discussion*

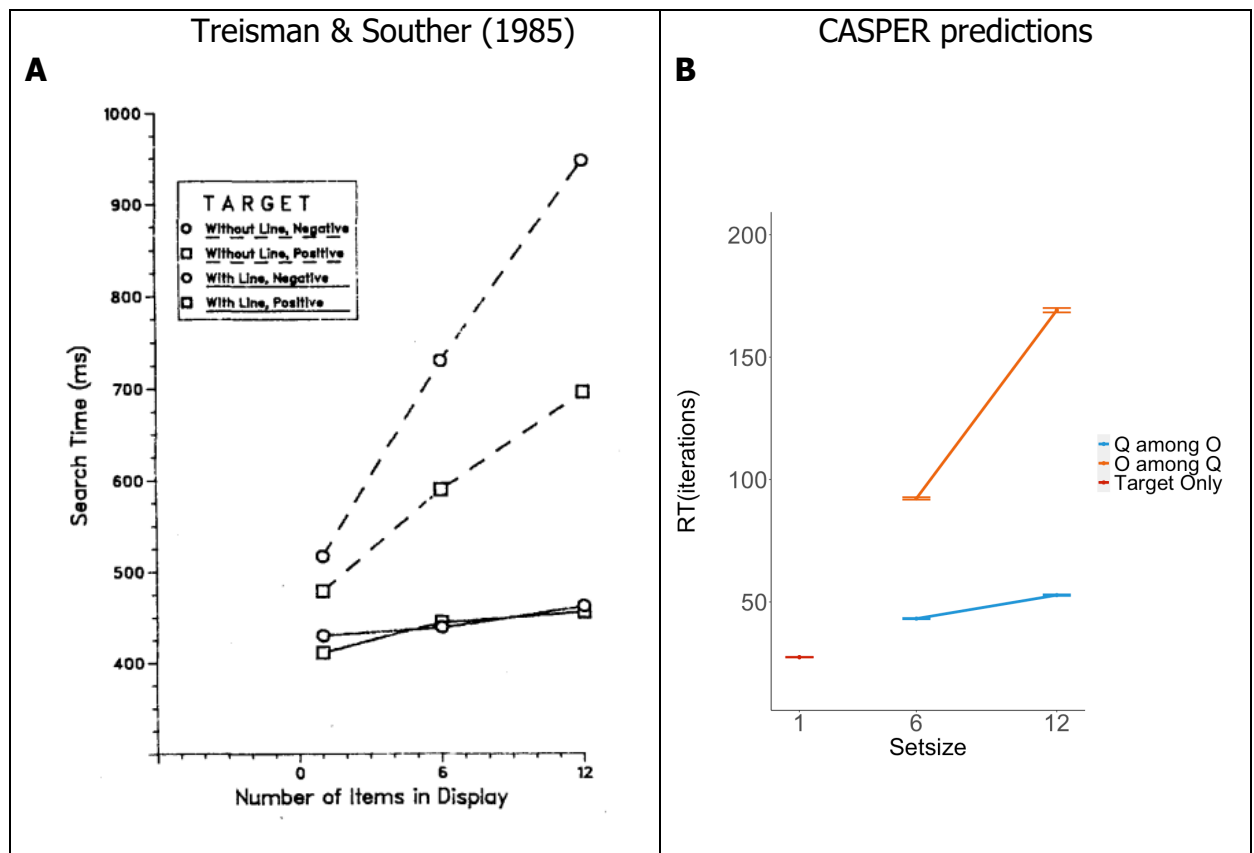

*Figure 3.4 A) Figure modified from Treisman & Souther (1985) to show only target-present searches. A search for an O among Qs is more difficult than a search for a Q among Os. B) CASPER's predictions for the same stimuli. CASPER predicts a search asymmetry but does not predict a difference in the target-only condition for different types of targets. Copyright © 1985 by the American Psychological Association. Reproduced with permission. Treisman, A., & Souther, J. (1985). Search asymmetry: A diagnostic for preattentive processing of separable features. Journal of Experimental Psychology: General, 114(3), 285–310.*

The mean over the virtual subject's mean response times per set size per condition is depicted in Figure 3.4B. The model showed a good qualitative fit to the empirical data, although there was a difference in the prediction of the target-only RT in the No Extra Feature condition (i.e., the target was an O), where CASPER underestimated the difficulty of the target-only condition as being similar to the difficulty of the target-only condition in the Extra Feature condition (where the target was a Q).

The coefficient of determination, $R^2$, was calculated between the empirical data and the simulation results each condition to assess fit. In this simulation, because there were only two non-target set sizes in the original experiment, the target-only condition was included in the calculation. For the No Extra Feature condition, $R^2 = 0.76$. For the Extra Feature condition, $R^2 = 0.98$. The combined total fit for the whole experiment, obtained by concatenating the simulation data from the two conditions into a single vector and comparing it with similarly concatenated human data, was $R^2 = 0.83$.

The estimated number of milliseconds per simulation iteration was computed for each condition by dividing the range of the human data within each condition by the range of the model data within each condition, including the target-only condition. Each simulation iteration in CASPER corresponds to 0.26ms and 1.04ms for the No Extra Feature and Extra Feature conditions, respectively.

In Simulation 4, the asymmetry between the two search conditions obtained due to the higher emphasis in CASPER's parallel processing algorithm on matching relevant values that are present in the search template, as compared to values that are present in a search item but not the search template. If a feature is present in the search template, its match or mismatch to a search item counts 30 times as much as it would if the feature were not present in the search template. Because this emphasis on the presence of the item in the search template is part of CASPER's parallel prioritization process, the account is consistent with Treisman & Souther's claim that preattentive processing of features present in the search template gives rise to search asymmetries.

CASPER overestimates the inefficiency of the more difficult condition (O among Qs) relative to the easier condition (Q among Os). CASPER also does not capture the asymmetry of the difficulty of the target-only condition. It is possible that since Treisman & Souther (1985) ran both target-present and target-absent trials within both conditions a response selection effect is responsible for the difference in target-only RT. CASPER does not have a mechanism to account for response selection effects that might influence the difficulty of target-only trials.

**Simulation 5: Configural effects**

Pomerantz et al. (1977) investigated the role of configural context in the search for visual stimuli. They found that the configuration of individual elements into more complex structures can either help or hinder the discriminability of the target from the distractors, as compared to simple features in isolation. In Pomerantz et al. (1977) Experiment 5, subjects performed searches in conditions that were designed to manipulate the helpfulness of the configuration to search. In the None condition, subjects searched for a difference among stimuli (e.g., a target 45-degree oriented line among oppositely oriented distractor lines; see Figure 3.5A). In the Good Context condition, extra vertical and horizontal lines in the shape of an L were added to the diagonally oriented lines from the None condition. This addition transformed the stimuli into diagonally oriented arrows and right triangles.[9] In the Poor Context condition, two vertical lines and one horizontal line were added to the stimuli in the None condition, creating abstract line shapes where the target was a left-right reflection of the distractors. This left-right reflection property is notable because Biederman & Cooper (1991b) found evidence of representational invariance for left-right reflections of compositions of parts into higher-order structural arrangements.

Thus Pomerantz et al.'s Good and Poor Context conditions can be understood as a search over emergent higher-order features instead of lower-order features (where the None condition does not include any higher-order features). According to CASPER's sixth claim that the visual system finds entry level representations most salient by default when available, the inclusion of higher-order object features to the Good and Poor Context conditions could account for the configural effects observed in their experiment (an explanation broadly consistent with Pomerantz et al.'s interpretation).

Pomerantz et al. found that in the None and Good conditions, RT was effectively flat over all set sizes, indicating parallel processing. However, the Poor Context search condition produced a steep search slope of 76ms/item, consistent with an extremely difficult search.

---

[9] See also Treisman & Souther (1985) Experiment 4

*Methods*

In Pomerantz et al. (1977) Experiment 5, subjects searched for a difference among the search items, rather than a specific target. As it is currently implemented, CASPER needs a target template, so for Simulation 5 the target template was given to CASPER as the item in the display that differed from the others in the original experiment (See Figure 3.5A). To simulate the findings in Pomerantz et al.'s experiments, CASPER's representations were extended to include representations of a generic arrow and a generic triangle (both representations are assumed to be invariant with left-right reflection). In the None condition, the search target was a 45-degree ($\pi/4$) diagonal, and the distractors were oppositely oriented ($3\pi/4$) diagonals. In the Poor context condition, the representation of the target shape included the low-level shape feature vector P1 (which corresponds to a horizontal line, two vertical lines, and the $\pi/4$ diagonal; see Table 2.3 for the encodings of the representations used in this simulation), concatenated with a vector of extra feature dimensions for the generic triangle. The representation of the distractors was composed of the low-level shape feature vector P2 (which differs from P1 only in the orientation of the diagonal) concatenated with the representation of the triangle.

In the Good context condition, the representation of the target was composed of the low-level shape feature vector G1 (which corresponds to a horizontal line, a vertical line, and a $\pi/4$ diagonal) concatenated with a vector of extra features representing the higher-order generic arrow. The representation of the distractors was composed of low-level shape feature vector G2 (which differed from G1 only in the orientation of the diagonal) concatenated with the vector representing the generic triangle. The higher-order vector representations of the arrow and the triangle were completely nonoverlapping, as defined in Table 3.2. The number of vector elements in the higher-order representations was explored during test simulations to determine an appropriate weight ratio of higher-order features to lower-order features during parallel processing, at ratios ranging from 16:1 to 128:1.

The model was given set sizes 1 (target-only), 2, 4, and 6, corresponding to the set sizes from the original experiment. Each simulation consisted of 100 trials per set size per condition. The model was run 64 times (i.e., for 64 full simulations), corresponding to 64 virtual subjects. Response times in units of model iterations were recorded.

|  | **Units 0...n-1** | **Units n...2n-1** |
|---|---|---|
| **Generic Arrow** | 000...0 | 111...1 |
| **Generic triangle** | 111...1 | 000...0 |

*Table 3.2 The vector representations for an arrow and a left-right invariant triangle. The representations are nonoverlapping. The length of the vectors was varied to investigate different relative strengths of the higher-order representations to the lower-order representations, where n could equal 8, 16, 32, 64, or 128. More units of representation approximate a greater representative strength in the CASPER model.*

*Results and Discussion*

The simulation data were analyzed in the same manner as the previous simulations. CASPER produced a good qualitative match to the results of the original experiment, where the None and Good Context conditions gave rise to efficient searches, while the plot of response time vs set size in the Poor Context condition had a steep slope characteristic of an inefficient search.

| Strength ratio of higher-order representations to lower-order representations | None Condition | Good Condition | Poor Condition | Model total |
|---|---|---|---|---|
| **32:1** | $R^2 = 0.98245$<br><br>2.59ms/ iteration | $R^2 = 0.99832$<br><br>3.38ms/ iteration | $R^2 = 0.99999$<br><br>10.22ms/ iteration | $R^2 = 0.927$ |
| **64:1** | $R^2 = 0.994$<br><br>2.51ms/ iteration | $R^2 = 0.988$<br><br>3.8ms/ iteration | $R^2 = 0.999$<br><br>8.04ms/ iteration | $R^2 = 0.938$ |
| **128:1** | $R^2 = 0.98702$<br><br>2.42ms/ iteration | $R^2 = 0.97774$<br><br>4.25ms/ iteration | $R^2 = 0.99995$<br><br>6.72ms/ iteration | $R^2 = 0.935$ |
| **256:1** | $R^2 = 0.99978$<br><br>2.82ms/ iteration | $R^2 = 0.99243$<br><br>3.68ms/ iteration | $R^2 = 0.99996$<br><br>6.2ms/ iteration | $R^2 = 0.932$ |

*Table 3.3 Model fits and estimated milliseconds per iteration for each condition and for the model overall, for each ratio of representational strength of higher-order visual features to lower-order visual features.*

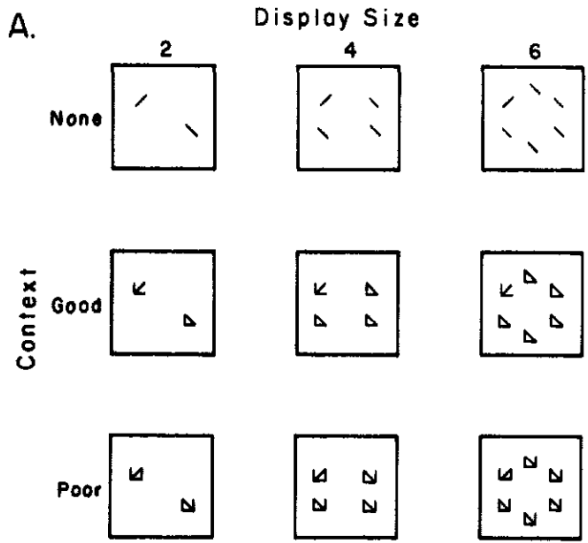

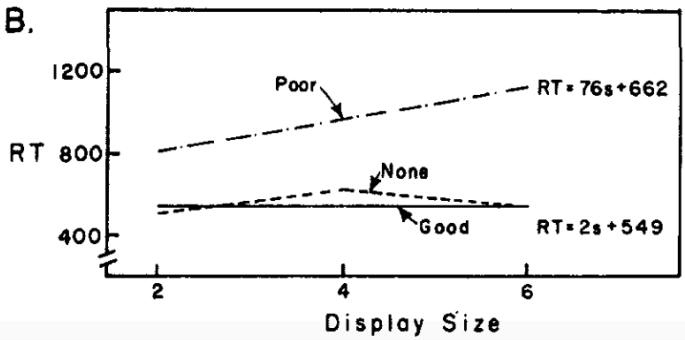

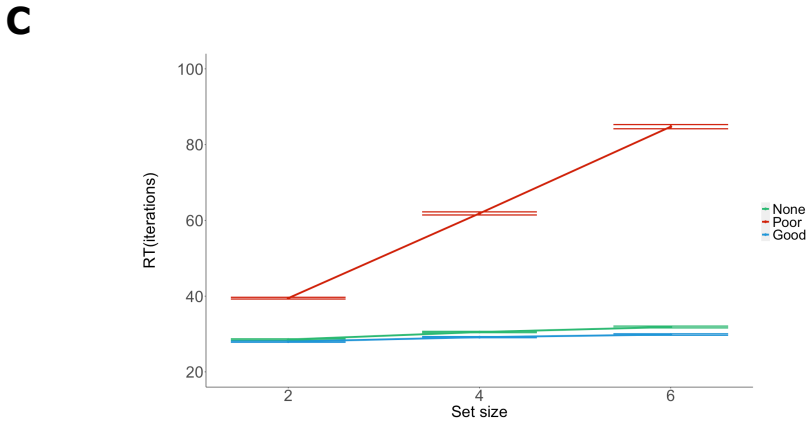

*Figure 3.5 A) Figure reproduced from Pomerantz et al. (1977) Experiment 5, showing Good, Poor, and None context condition displays. B) Pomerantz et al. showed that the Good Context and None conditions produced efficient searches with shallow search slopes while the Poor Context condition produced a steep search slope, indicating that poor configural effects dramatically increase the difficulty of search. C) CASPER's simulation of Pomerantz et al. (1977) produces the same qualitative pattern of results as the original experiment for all weight ratios of higher-order vs lower-order representational strength, including 128:1 pictured here. The Good context and None condition are efficient searches, while the Poor Context condition produces a steep linear search slope. Copyright © 1977 by the American Psychological*

*Figure 3.5 (cont.) Association. Reproduced with permission. Pomerantz, J. R., Sager, L. C., & Stoever, R. J. (1977). Perception of wholes and of their component parts: Some configural superiority effects. Journal of Experimental Psychology: Human Perception and Performance, 3(3), 422–435.*

The coefficient of determination, $R^2$, was calculated for each condition and across conditions for each ratio of higher- to lower-order representational weight. The best fit across both milliseconds per iteration and $R^2$ was obtained from a ratio of 128:1 (depicted in Figure 3.5C). The model fits for the various weight ratios and the number of milliseconds per simulation iteration per condition are reported in Table 3.3, along with estimates of the number of milliseconds in empirical data accounted for by the model per condition per weight ratio. Note that the improvement in alignment of milliseconds per iteration across conditions from 32:1 to 64:1, etc., produced diminishing returns as the weight ratio was increased, suggesting that there exists a point at which the lower-order representations are functionally overwhelmed by higher-order representations and additional weight on the higher-order representations isn't helpful.

In Simulation 5, the higher-order generic representations of triangles and arrows were added to existing lower-order feature representations, with greater representational weight (i.e., greater ratios of higher-order to lower-order representational units). This change caused CASPER to put greater emphasis on the higher-order representations in the similarity match that is computed as part of the parallel processing routine that determines selection priority. In the Good Contrast condition, this additional level of representation transforms the search into an easy contrast between a target and distractor that do not overlap in feature space (i.e., the generic arrow and generic triangle). In the Poor Contrast condition, the search is transformed into a difficult search where most of the representational content is completely overlapping (i.e., the generic triangle), with only a small amount of representational weight dedicated to the lower-order differences between target and distractors.

That greater representational weights on higher-order features compared to lower-order features achieved better fits to the empirical data is consistent with CASPER's theoretical claim that the visual system places higher priority on entry levels of representation (see also Enns & Rensink, 1991).

However, as in other simulations, CASPER still underestimates the difficulty difference between easy and difficult searches in this simulation. It is unclear whether this underestimate of the relative difference in search difficulty is due to a representation or process inherent to visual search that CASPER is not capturing, or whether a longer or more difficult search process prolongs search time for reasons that lie outside of the task itself (i.e., momentary lapses in attention on the part of subjects).

# Chapter 4: Predicting combinations of relations and features in search with CASPER

**Background**

Pomerantz et al. (1977) demonstrated the impact of higher-order representations on visual search (see also Enns & Rensink, 1991). In their experiments, a 'good' configural arrangement was consistent with a higher-order representation and facilitated search, whereas a 'poor' configural arrangement was inconsistent with a meaningful higher-order arrangement and did not facilitate search. The higher-order representations that provided configural context in these experiments included axes of symmetry and simple bounded regions (surfaces) that are likely to reside in V4 (Vannuscorps et al., 2021). More sophisticated representations of shape appear to involve even later areas in ventral processing, including LOC (Goodale et al., 1991). Another important function of LOC is to compute the spatial relations between objects in a scene and between parts of an object (Kim & Biederman, 2011).

Search for relations is also informative with respect to the role of binding in visual processing. Although search for feature conjunctions has generally been taken to require visual binding under attention (e.g., Treisman & Gelade, 1980), search for conjunctions is known to get faster with practice (e.g., over successive blocks of trials). This effect of practice suggests that such conjunctive bindings can become automatized (Logan, 1994). By contrast, Logan (1994) showed that search for visual relations produces a steep and linear search slope that does not get faster with practice, suggesting that binding objects (e.g., a plus or a dash as in Logan's experiments) to relational roles (e.g., above and below) cannot be automatized (however, see Clevenger, 2017, for evidence that repeating a relational target across experiment trials may assist in the process of binding in relations by offloading working memory demands to long-term memory).

**Extending CASPER to represent relations**

The representations in CASPER can be extended to bind relational roles to fillers within search items. To do this, CASPER represents a search item as *n* vectors, where each vector captures one of the roles in the relation (e.g., for a relation such as *above* (*x*, *y*), *n*=2, where the first vector corresponds to the representation of *above+x*, and the second vector corresponds to the representation of *below+y*). To represent a target consisting of a red X above a green O, the vector representations for *red*, *X*, and *above* are concatenated to represent *above+red+X*, and the vector representations for *green*, *O*, and *below* are concatenated to represent *below+green+O.* The resulting pair of vectors represents the complete target template. Analogous pairs of vectors represent the search items. If relations can be processed pre-attentively, then the relation information would be represented in the search items prior to attentional selection; but if not, then the relation information would not be part of the search items prior to selection and would instead be calculated at the time of attentional selection.

Given the evidence that the binding of relational roles to their arguments requires attention, CASPER ignores relational roles during parallel processing. The parallel processing routine only looks at color and shape features, rather than the presence of the color and shape features in the proper relational roles. For example, for a target consisting of a red X over a green O, CASPER would compute the same degree of similarity for the target as it would for a green O over a red X or a green X over a red O. In this example, in which only the relational roles are swapped, CASPER predicts that parallel processing will fail to discriminate the target from the distractors. Therefore, search for relations should be serial, and RT should be steep and linear as a function of set size.

**Simulation 6: Search for combinations of features and relations**

Simulation 6 was designed to test CASPER's extended representation scheme across different combinations of features and relations for its ability to make new predictions. The simulation consisted of three conditions: Relation-only, Feature-only, and relation+feature. The expectation, based on the previous literature and prior simulations, is that CASPER will predict a steep linear function of RT vs set size in the Relation-only condition (as obtained by Logan, 1994; see Figure 4.1A) and a steep but negatively accelerating relationship of RT to set size in a difficult Feature-only condition (as in Buetti et al., 2016). The effect of combining feature and relation information on search performance is less intuitive without running the model.

*Methods*

The target in each condition was a red X over a green O, represented as described in the example above, using the same color and shape vectors used in Simulations 1 through 5 (see Tables 2.2 and 2.3). The search items for each condition are shown in Figure 4.1. In the Feature-only condition, the distractors differed from the target in that the red X was replaced with an orange X. The feature-based difference along the color dimension was designed to be comparable to the difference in color for Simulation 3 that modeled the condition from Buetti et al. (2016) Experiment 1A, where the distractors were orange and the search target was red (i.e., a difficult feature-based search). In the Relation-only condition, the bindings of the relational roles to the fillers were reversed, such that the distractors were green Os above red Xs. Otherwise, there were no differences between the features of the target and distractors. This condition is the most directly analogous to Logan's (1994) experiments, although it does include (non-diagnostic) color information that was not present in Logan's original experiments. In the relation+feature condition, the red X was replaced by the orange X, and the relational roles of the Xs and Os were reversed (as in the Relation-only condition) so that the distractors were green Os over orange Xs.

The model was given set sizes 1 (target-only), 2, 4, 8, and 16, with items arranged radially around the center of the model's visual field. Each simulation consisted of 52 trials per set size per condition. As in previous simulations, 32 full simulations were run, corresponding to 32 virtual subjects. Response times in units of model iterations were recorded.

| Target | Relation-only | Feature-only | Relation + feature |
|---|---|---|---|
| 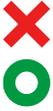 | 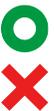 | 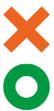 | 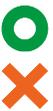 |

*Figure 4.1 Search stimuli for simulation 6. The target was always a red X over a green O. In the Relation-only condition, the distractors were a green O over a red X, in a reversal of the relational roles to their fillers. In the Feature-only condition, the distractors differed from the target in the hue of the X, which was orange instead of red. In the Relation + feature, the relational roles of the X and O were swapped, and the X also differed in hue from the target in the way that it did in the Feature-only condition.*

*Results and Discussion*

The simulation results are shown in Figure 4.2B. As expected, for the Relation-only condition, CASPER predicts that the search function will be steep and linear as observed by Logan (1994; Figure 4.2A). For set size 8 (the largest in Logan's experiment), the mean response time in model iterations was 113.23 iterations. For set size 2, the mean response time was 41.09 iterations. CASPER therefore predicts a search slope of 12.02 iterations per search item in the Relation-only condition. Logan determined that for the above-below relations, the search slope was 85ms/item for the target-present condition. Therefore, in the Relation-only condition, each iteration in CASPER corresponds to 7.07ms (compare this estimate to 6.3ms/iteration for Simulation 1 of Treisman & Gelade, 1980, and 6.72ms/iteration for the Poor condition in Simulation 5 of Pomerantz et al., 1977).

CASPER predicts that both Feature-only and relation+feature searches will proceed similarly, with negatively accelerating functions of RT vs set size, indicating more parallel processing. Feature information appears to provide enough contrast between the target and distractors during CASPER's parallel processing to make the resulting search effectively equivalent to a difficult feature-based search. CASPER predicts no added benefit of relational information when features alone would predict efficient search.

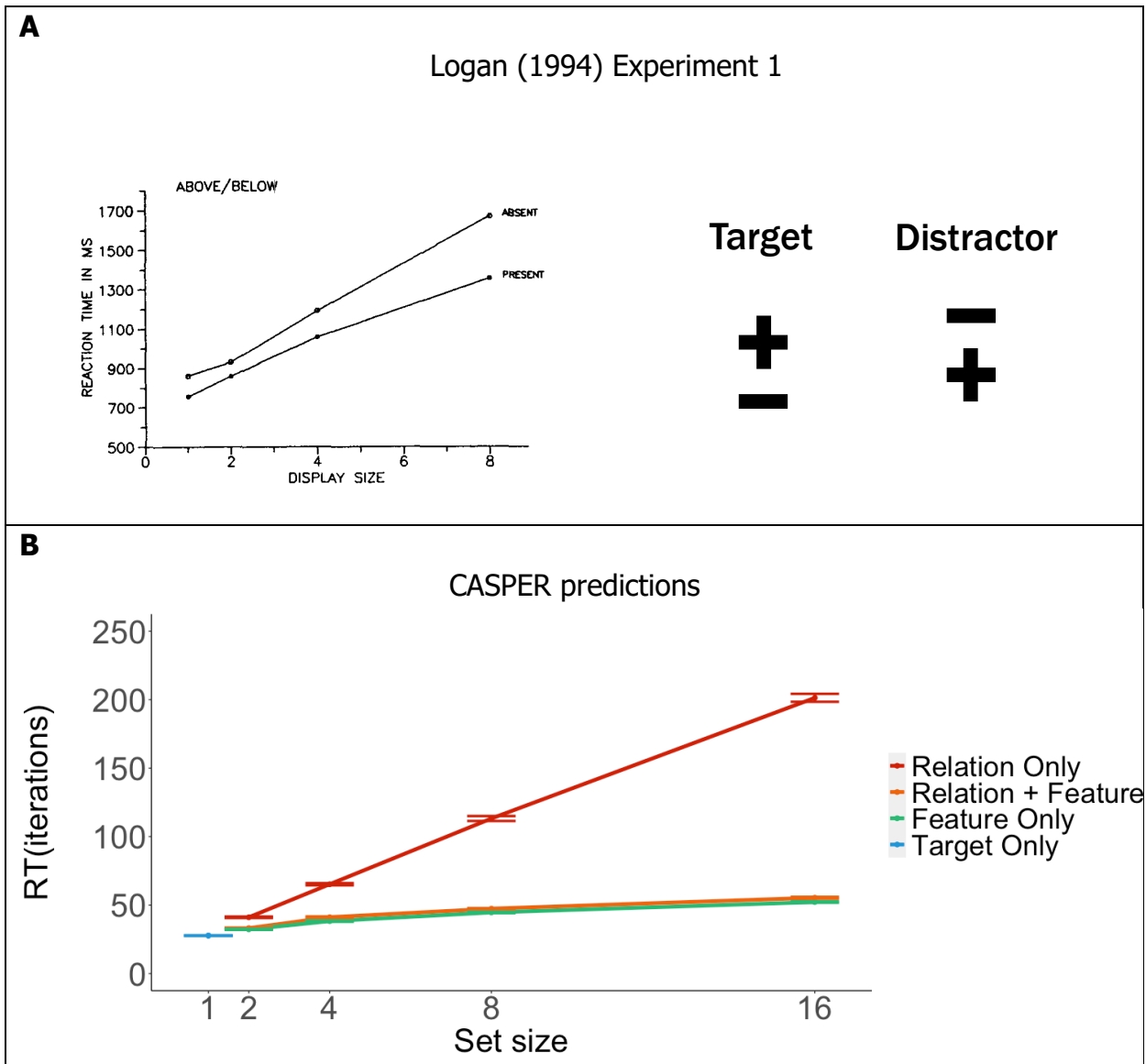

*Figure 4.2 A) Figure from Logan (1994) Experiment 1, showing the function of reaction/response time (RT) vs set size for a search among items defined by plus and dash shapes in above-below relations. B) CASPER's prediction for a Relation-only search (similar to Logan's (1994) experiments, as well as for a Feature-only (color) search, and a feature+ relation search where both feature and relation information discriminate the target from the distractors. Copyright © 1994 by the American Psychological Association. Reproduced with permission. Logan, G. D. (1994). Spatial attention and the apprehension of spatial relations. Journal of Experimental Psychology: Human Perception and Performance, 20(5), 1015–1036.*

**Simulation 7: Limiting search items to one color**

Simulation 6 used stimuli that were unlike Logan's (1994) stimuli in that there were multiple color features present within a search item. Simulation 7 limited the stimuli to one color per search item.

*Methods*

The target in each condition was a red X over a red O and was represented using the same color and shape vectors used in Simulations 1 through 6 (see Tables 2.2 and 2.3). In the Feature-only condition, the distractors differed from the target in that the color red was replaced with the color orange. In the Relation-only condition, the bindings of the relational roles to the fillers were reversed, such that the distractors were red Os above red Xs. This condition is even more directly analogous to Logan's (1994) experiments, although Logan used plus and dash shapes instead of X and O shapes. In the relation+feature condition, both the colors and the relational roles within the distractors were changed, so that the distractors were orange Os over orange Xs. The search items for each condition are shown in Figure 4.3.

As in Simulation 6, the model was given set sizes 1 (target-only), 2, 4, 8, and 16, with items arranged radially around the center of the model's visual field. Each simulation consisted of 52 trials per set size per condition. As in previous simulations, 32 full simulations were run, corresponding to 32 virtual subjects. Response times in units of model iterations were recorded.

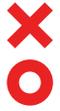

*Figure 4.3 Search stimuli for simulation 6. The target was always a red X over a red O. In the Relation-only condition, the distractors were a red O over a red X, in a reversal of the relational roles to their fillers. In the Feature-only condition, the distractors varied from the target in the hue of the X and O, which was orange instead of red. In the Relation + feature, the relational roles of the X and O were swapped, and the hue of the X and O differed from the target in the way that it did in the Feature-only condition.*

*Results and Discussion*

The simulation results are shown in Figure 4.4. CASPER's predictions for searches using single-color items were qualitatively the same as the predictions with dual-color items. In the Relation-only search condition, the slope was calculated in the same way as in Simulation 6, and it was similarly steep at a predicted 13.25 simulated iterations per search item. This corresponds to 6.42ms per simulation iteration in CASPER, which is close to the estimate of 7.07ms in Simulation 6. The Feature-only and Relation+feature conditions again produced very similar results.

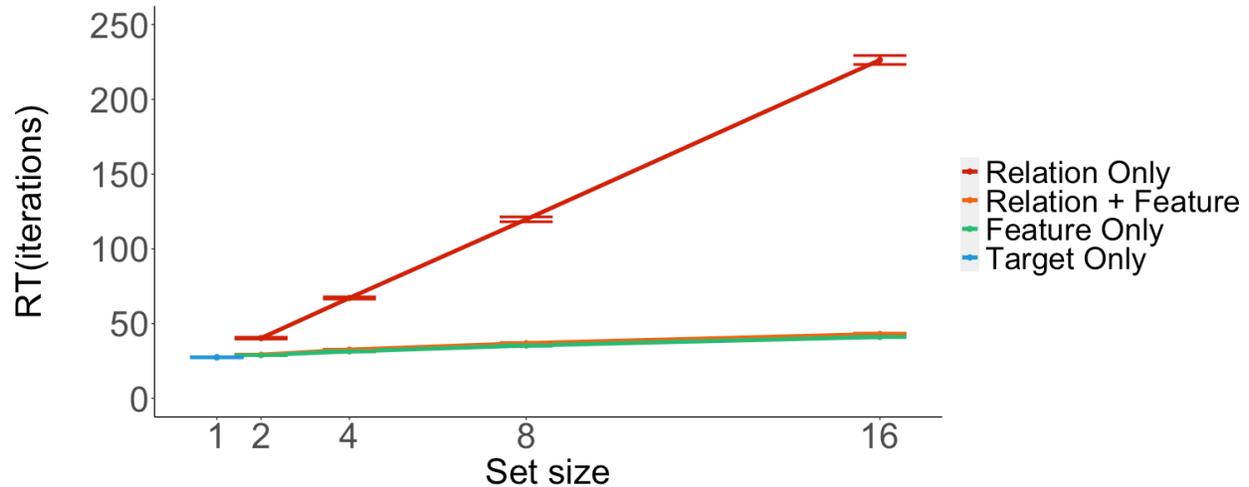

*Figure 4.4 CASPER's predictions Simulation 7 using single-color search items for Relation-only, Feature-only, and relation+feature searches. CASPER predicts that search for only relational information is steep and linear, while the conditions containing feature information are predicted to benefit from parallel processing.*

**Summary**

Consistent with Logan (1994), CASPER predicts steep linear search slopes when only relational information is present. In the simulations in this chapter, CASPER also did not predict any sort of additive benefit of the presence of both feature and relation information in search. Relation + feature simulations produced the same result as Feature-only conditions because CASPER does not use relational information in parallel processing. As in earlier simulations, CASPER also predicts that the more feature dimensions the target and the distractors have in common, the more difficult the search becomes. In the current simulations, this effect manifested in increased difficulty for feature searches with multicolor items.

# Chapter 5: Experiments testing CASPER's predictions

**Introduction**

As described in the previous chapter, CASPER predicts steep linear functions of response time to set size as Logan (1994) found in experiments using relations. CASPER further predicts that there will be no added benefit of relational information to search efficiency when feature information is present, with no difference between Feature-only and relation+feature searches. In seven experiments (four main experiments and three replications) CASPER's predictions for combinations of features and relations in search displays were tested on human participants.

**General Methods**

All experiments were run online on the Pavlovia web platform during the COVID-19 pandemic. The experiments were written in Javascript using the PsychoJS toolbox. The displays were designed to be presented at a constant size regardless of the computer display the participant used to participate in the experiment. At the beginning of the experiment, participants were asked to use arrow keys to adjust an image of a credit card to match the size of a real credit card; the amount of adjustment that was recorded was used to scale the presentation of the stimuli to the display.

Each trial began with a fixation cross presented in the center of the display for 500ms. Search items were arranged radially in two concentric circles around the central fixation location. The diameter of the inner circle was 5.95cm and the diameter of the outer circle was 11cm. There were 12 possible locations in the inner circle and twelve possible locations in the outer circle, ranging from 0 radians to $11\pi/12$ radians, separated by $\pi/12$ radians. Up to 0.25cm of jitter was applied in each direction to each location in the display. Search items were displayed such that they were 1 cm in height. Example displays can be found in Figures 5.1, 5.3, 5.5, and 5.8.

In all experiments, there was one search target present in a display. Participants were asked to indicate, using the left or right arrow key, on which side of the target a small black dot appeared. Response time and accuracy were recorded for each trial. Participants were given practice trials before beginning experimental trials. Each experiment contained 12 blocks of 52 trials, for 624 total trials per experiment.

During each trial, the display was presented for up to 8 seconds before a timeout would end the trial and be recorded as an incorrect response. At the end of a trial, participants were given feedback that their response was either correct or wrong. This feedback lasted 500ms, followed by a 500ms pause between trials.

Participants were removed if they achieved low accuracy (less than 90%). Individual trials were removed if the response time was faster than 300ms or if the trial response was incorrect.

**Experiments 1a & 1b**

Experiments 1a and 1b investigated the contributions of features, relations, and the combination of features and relations to the detection of the search target in a display. In one condition the search task could be performed using a difficult feature search, where the difference in hue between the target and the distractors was 15 degrees (i.e., red vs. reddish orange). A second condition required the use of relations, which according to Logan (1994) Experiment 1 would require serial attentional scrutiny, resulting in a steeply linear search function. A third condition combined the difficult feature search and the relation search to see whether any benefit could be gained from the availability of both feature and relation information.

*Methods*

In both Experiments 1a and 1b, the target was always a red 'X' over a green 'O'. In the 'Feature-only' condition, the distractors were an orange 'X' over a green 'O'; the hue of the orange 'X' was 15 degrees separated from the hue of the red 'X' in the target. In the 'Relation-only' condition, the distractors were a green 'O' over a red 'X', reversing the relational roles of the 'X' and the 'O' compared to the target. In the 'Relation + feature' condition, the distractors were a green 'O' over an orange 'X' that was 15 degrees away in hue from the red 'X' in the target, as in the 'Feature-only' condition. There was also a 'Target Only' condition where only the target appeared in the display. Total display set sizes, including both the target and distractors, were 1, 2, 4, 8, and 16. The search items were presented at locations chosen randomly from the 24 possible locations within the radial display. Example displays from each condition in Experiments 1a and 1b are shown in Figure 5.1.

In Experiment 1a, in the 'Relation + feature' condition, black dots on the distractor items were only presented on one side. In Experiment 1b, this error was corrected, which served to both eliminate the possibility of an accidental confound and replicate the results of Experiment 1a.

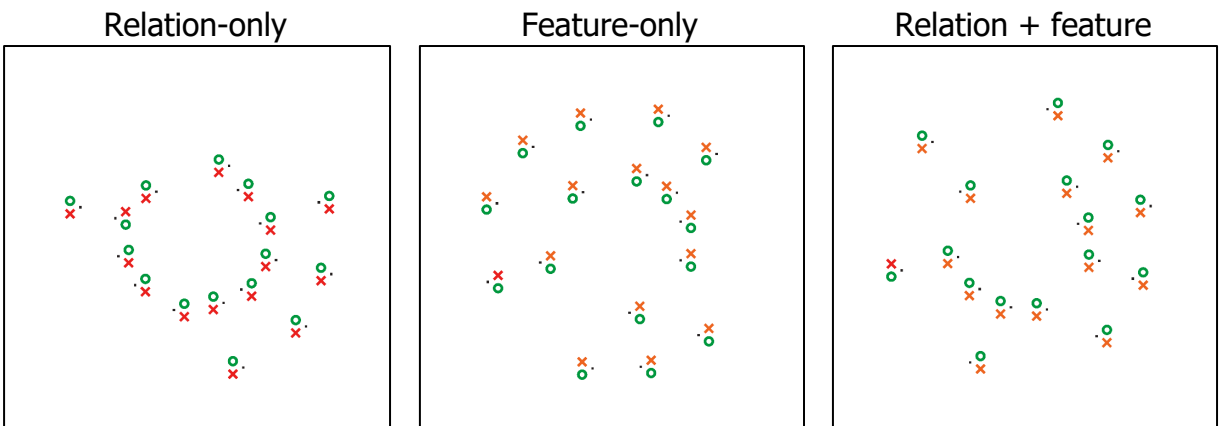

*Figure 5.1 Example displays for set size 16 for Experiments 1a & 1b*

The experimental and recruitment procedures were approved by the Institutional Review Board at the University of Illinois Urbana-Champaign. For Experiment 1a, 55 participants were recruited from the Department of Psychology Subject Pool at the University of Illinois Urbana-Champaign during October 2020. Seventeen participants were removed due to low accuracy (less than 90%), resulting in 38 included participants who were between 18-22 years old. For Experiment 1b, 63 new participants were recruited from the Department of Psychology Subject Pool at the University of Illinois Urbana-Champaign between May and September of 2021. Sixteen participants were removed due to low accuracy, resulting in the inclusion of 47 included participants who were between 18 and 34 years old.

*Results*

The results of Experiments 1a and 1b are shown in Figure 5.2. All conditions produced a negatively accelerating relationship of response time (RT) vs set size. A log slope was estimated per condition for each participant using a regression on log-transformed data, and the mean was computed across participants. For both experiments, the mean log slope for the 'Relation-only' condition was steeper than the mean log slopes for the 'Feature-only' and 'Relation + feature' conditions.

For Experiment 1a, the mean log slopes were 596ms/log unit, 251ms/log unit, and 195ms/log unit for the 'Relation-only', 'Feature-only', and 'Relation + feature' conditions, respectively. Using the estimated log slope per condition for each participant, paired t-tests were performed between all combinations of conditions. There was a statistically significant difference between the 'Relation-only' and each of the 'Feature-only' and 'Relation + feature' conditions (t = 18.937, df = 37, p < .001; t = 21.423, df = 37, p < .001, respectively). There was also a statistically significant difference between the 'Feature-only' and 'Relation + feature' conditions (t = -5.7011, df = 37, p < .001).

For Experiment 1b, the same analysis was performed as Experiment 1a. The mean log slopes were 574ms/log unit, 249ms/log unit, and 192ms/log unit for the 'Relation-only', 'Feature-only', and 'Relation + feature' conditions, respectively. There was a statistically significant difference between the 'Relation-only' condition and each of the 'Feature-only' and 'Relation + feature' conditions (t = 14.291, df = 46, p < .001; t = 17.556, df = 46, p < .001, respectively). There was also a statistically significant difference between the 'Feature-only' and 'Relation + feature' conditions (t = -5.3343, df = 46, p < .001).

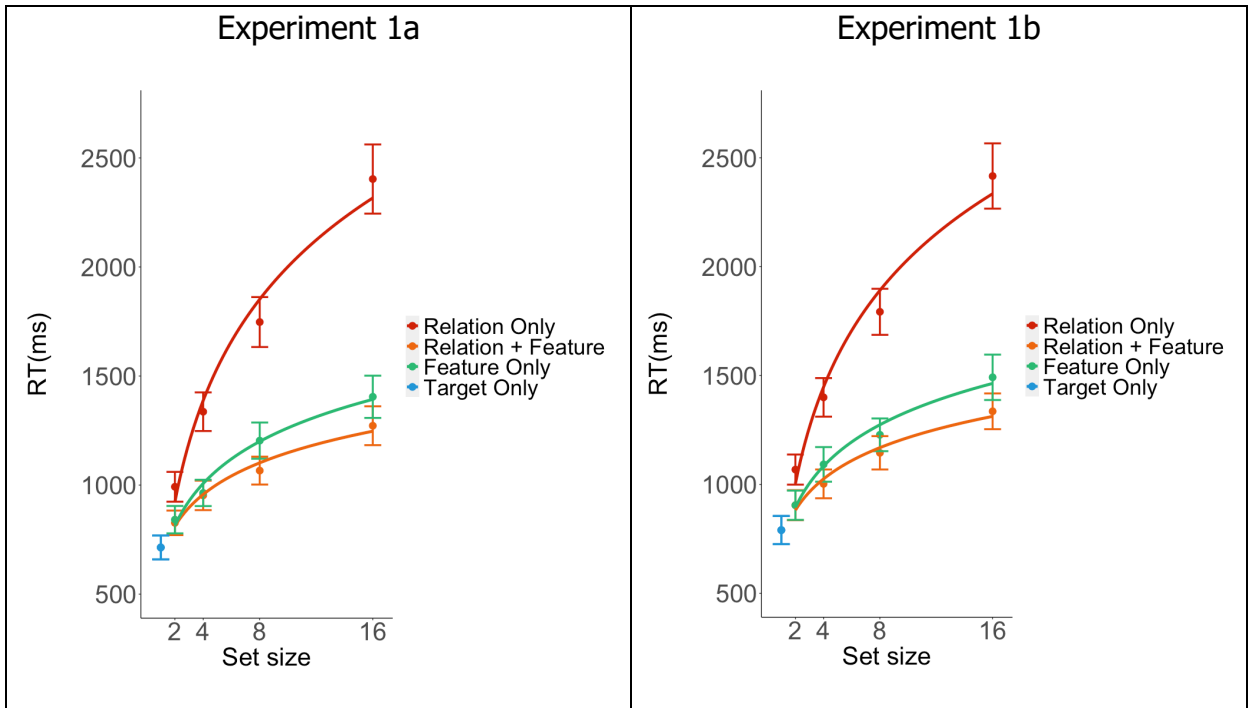

*Figure 5.2 Response time as a function of set size for Experiments 1a (left) and 1b (right). The 'Relation-only' condition is shown in red and the 'Feature-only' and 'Relation + feature' conditions are shown in green and orange, respectively. The 'Target Only' condition is shown in blue. The 'Relation-only' condition had a steeper log slope than the 'Feature-only' and 'Relation + feature' conditions. The log slopes were 596ms/log unit, 251ms/log unit, and 195ms/log unit for the 'Relation-only', 'Feature-only', and 'Relation + feature' conditions, respectively, for Experiment 1a. The log slopes were 574ms/log unit, 249ms/log unit, and 192ms/log unit for the 'Relation-only', 'Feature-only', and 'Relation + feature' conditions, respectively, for experiment 1b. There was a statistically significant difference among all combinations of conditions. Error bars indicate the standard error of the mean.*

## Discussion

Experiment 1b replicated the qualitative trends in Experiment 1a. In both experiments, all conditions resulted in a negatively accelerating relationship between response time and set size. Notably, the presence of a negatively accelerating curve in the 'Relation-only' condition suggests parallel processing was occurring in a putatively relational search. This is inconsistent with Logan's (1994) findings for relational search. However, the 'Relation-only' condition had a much larger log slope than either the 'Feature-only' condition or the 'Relation + feature' condition, which suggests that this is still a more difficult search than one where feature information can be used to identify the target—even a classically difficult feature-based search. There was also a statistically significant difference between the 'Feature-only' and the 'Relation + feature' condition, with evidence of more parallel processing in the 'Relation + feature' condition than the 'Feature-only' condition.

**Experiments 2a & 2b**

The results of Experiments 1a and 1b are surprising because they fail to replicate Logan's (1994) Experiment 1 result. However, Logan's search items were only one color. Experiments 2a and 2b investigated whether color information within a search item was providing some extra benefit to help participants find the search target. In these experiments, each search item was limited to only one color. This allowed for a conceptual replication of Logan's relation experiment since all the search items in the display were one color in the Relation-only condition.

*Methods*

In both Experiment 2a and 2b, the target was always a red 'X' over a red 'O'. In the 'Feature-only' condition, the distractors were an orange 'X' over an orange 'O', where the orange hue was 15 degrees separated from the hue of the target. In the 'Relation-only' condition, the distractors were a red 'O' over a red 'X', reversing the relational roles of the 'X' and the 'O' compared to the target. In the 'Relation + feature' condition, the distractors were an orange 'O' over an orange 'X' that was 15 degrees separated in hue from the red target. There was also a 'Target Only' condition where only the target appeared in the display. Total display set sizes, including both the target and distractors, were 1, 2, 4, 8, and 16. The search items were presented at locations chosen randomly from the 24 possible locations within the radial display. Example displays from each condition in Experiments 2a and 2b are shown in Figure 5.3.

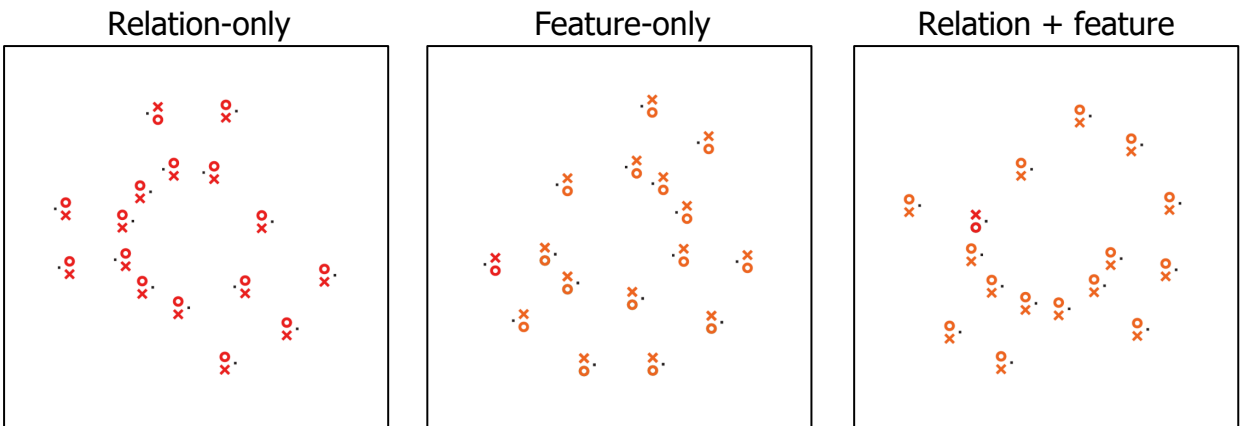

*Figure 5.3 Example displays for set size 16 for Experiments 2a & 2b*

In Experiment 2a, in the 'Relation + feature' condition, black dots on the distractor items were only presented on one side. In Experiment 2b, this error was corrected, which served to both eliminate the possibility of an accidental confound and replicate the results of Experiment 2a.

The experimental and recruitment procedures were approved by the Institutional Review Board at the University of Illinois Urbana-Champaign. For Experiment 2a, 35 participants were recruited from the Department of Psychology Subject Pool at the University of Illinois Urbana-Champaign during October and November 2020. Three participants were removed due to low accuracy (less than 90%), resulting in 32 included participants between 18 and 21 years old. For Experiment 2b, 44 new participants were recruited from the Department of Psychology Subject Pool at the University of Illinois Urbana-Champaign during September 2021. Six participants were removed due to low accuracy, resulting in the inclusion of 38 included participants between 18 and 28 years old.

*Results*

The results of Experiments 2a and 2b are shown in Figure 5.4. All conditions produced a negatively accelerating function of response time (RT) vs set size. A log slope was estimated per condition for each participant using a regression on log-transformed data, and the mean was computed across participants. For both experiments, the log slope for the 'Relation-only' condition was steeper than the log slopes for the 'Feature-only' and 'Relation + feature' conditions.

For Experiment 2a, the mean log slopes were 571ms/log unit, 139ms/log unit, and 115ms/log unit for the 'Relation-only', 'Feature-only', and 'Relation + feature' conditions, respectively. The estimated log slope per condition for each participant was used to perform paired t-tests between all combinations of conditions. There was a statistically significant difference between the 'Relation-only' and each of the 'Feature-only' and 'Relation + feature' conditions ($t(31) = 17.005$, $p < .001$; $t(31) = 15.109$, $p < .001$, respectively). There was not a statistically significant difference between the 'Feature-only' and 'Relation + feature' conditions ($t(31) = -1.1639$, $p = .2534$)

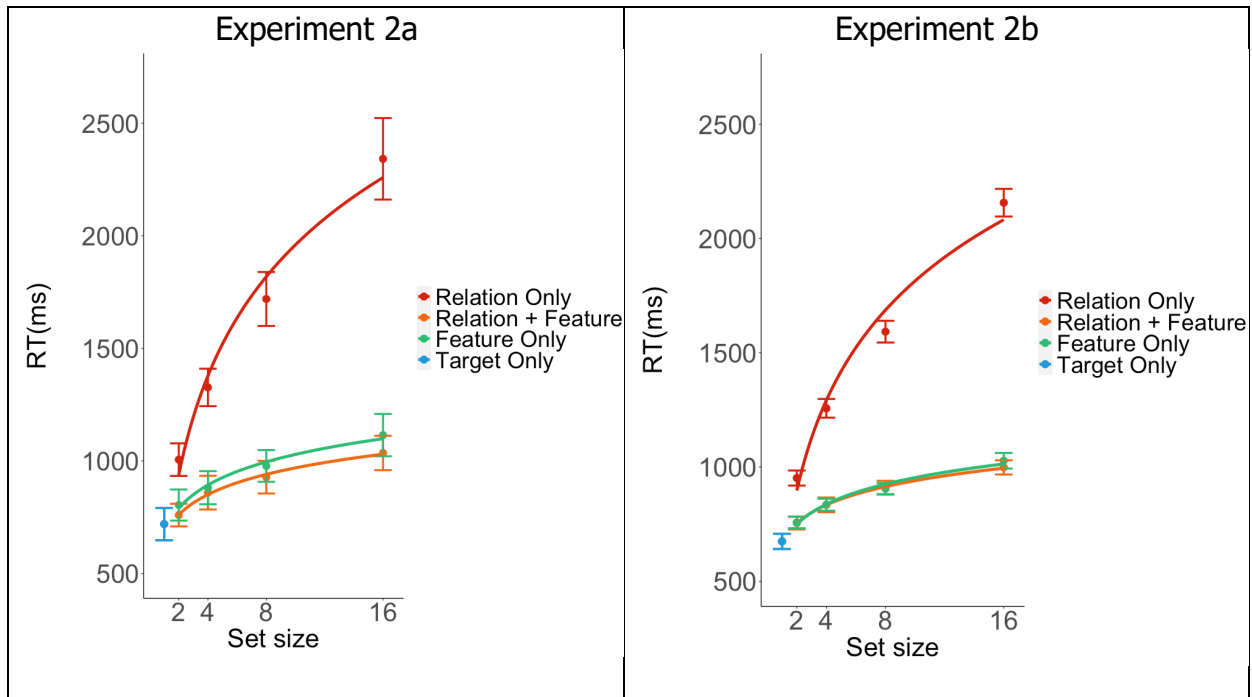

*Figure 5.4 Response time as a function of set size for Experiments 2a (left) and 2b (right). The 'Relation-only' condition is shown in red and the 'Feature-only' and 'Relation + feature' conditions are shown in green and orange, respectively. The 'Target Only' condition is shown in blue. The 'Relation-only' condition had a steeper log slope than the 'Feature-only' and 'Relation + feature' conditions. There was not a statistically significant difference between the 'Feature-only' and 'Relation + feature' conditions. Error bars indicate the standard error of the mean. For Experiment 2a, the mean log slopes were 571ms/log unit, 139ms/log unit, and 115ms/log unit for the 'Relation-only', 'Feature-only', and 'Relation + feature' conditions, respectively. The mean log slopes for Experiment 2b were 520ms/log unit, 123ms/log unit, and 115ms/log unit for the 'Relation-only', 'Feature-only', and 'Relation + feature' conditions, respectively.*

For Experiment 2b, the same analysis was performed as in Experiment 2a. The mean log slopes were 520ms/log unit, 123ms/log unit, and 115ms/log unit for the 'Relation-only', 'Feature-only', and 'Relation + feature' conditions, respectively. There was a statistically significant difference between the 'Relation-only' condition and each of the 'Feature-only' and 'Relation + feature' conditions, $t(37) = 24.186$, $p < .001$; $t(37) = 24.426$, $p < .001$, respectively. There was not a statistically significant difference between the 'Feature-only' and 'Relation + feature' conditions, $t(37) = -1.1208$, $p = .2696$).

*Discussion*

Experiment 2b replicated the qualitative trends in Experiment 2a. In both experiments, all conditions resulted in a negatively accelerating relationship between response time and set size, indicating that there was parallel processing occurring during search. Again the 'Relation-only' condition had a much larger log slope than either the 'Feature-only' condition or the 'Relation + feature' condition, which suggests that this is still a more difficult search than one where feature information can be used to identify the target, even when each search item is just one color. Unlike Experiments 1a and 1b, however, there was not a statistically significant difference between the 'Feature-only' and the 'Relation + feature' condition.

These results suggest that color information was not giving subjects an advantage in relational search. Indeed, multiple colors within a stimulus as in Experiments 1a and 1b may have made search more difficult.

**Experiment 3**

The arrangement of an 'O' above an 'X' has a structural configuration similar to a person. The extrastriate body area (EBA) and fusiform body area (FBA) are specialized regions of the brain that have been demonstrated to selectively process body parts and whole-body figures (Downing et al., 2001; Taylor et al., 2007). Rapid or holistic processing of the shapes of human figures and body parts by the visual system is likely adaptive for human survival. If the visual system is processing the stimuli as body figures instead of relational stimuli, this could explain parallel processing. In Experiment 3, the 'heads' of the search items (Os) were replaced with squares to diminish the likelihood of a boost from brain areas that are specialized to process bodies.

*Methods*

Experiment 3 used the same methods as Experiment 2, but the O shapes were replaced by squares. Example displays from each condition in Experiment 3 are shown in Figure 5.5.

The experimental and recruitment procedures were approved by the Institutional Review Board at the University of Illinois Urbana-Champaign. Forty-seven participants were recruited from the Department of Psychology Subject Pool at the University of Illinois Urbana-Champaign during October and November 2020. Ten participants were removed due to low accuracy (less than 90%), resulting in 37 included participants aged between 18 and 27 years old.

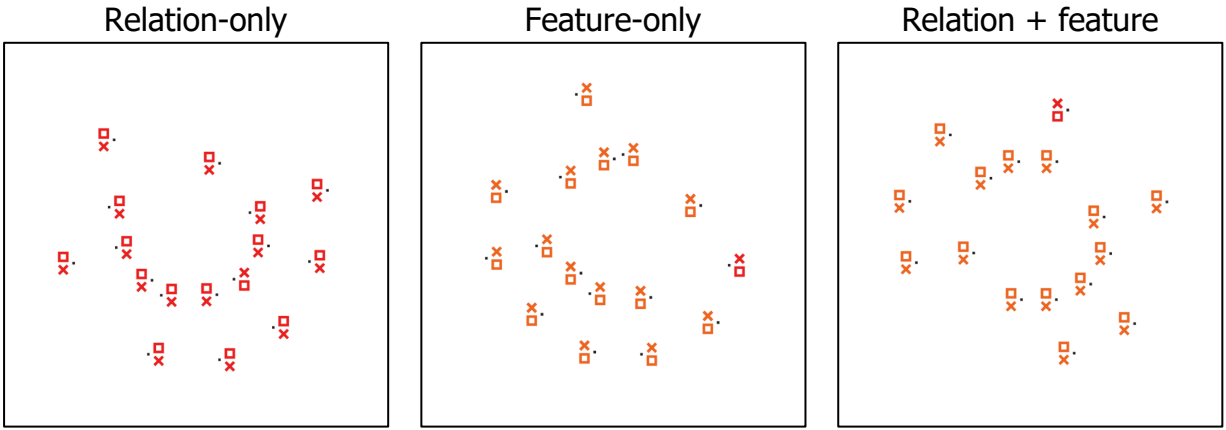
*Figure 5.5 Example displays for set size 16 for Experiment 3*

*Results*

The results of Experiment 3 are shown in Figure 5.6. All conditions once again produced a negatively accelerating function of response time (RT) vs set size. A log slope was estimated per condition for each participant using a regression on log-transformed data, and the mean was computed across participants. For both experiments, the log slope for the 'Relation-only' condition was steeper than the log slopes for the 'Feature-only' and 'Relation + feature' conditions.

For Experiment 3, the mean log slopes were 536ms/log unit, 126ms/log unit, and 112ms/log unit for the 'Relation-only', 'Feature-only', and 'Relation + feature' conditions, respectively. The estimated log slope per condition for each participant was used to perform paired t-tests between all combinations of conditions. There was a statistically significant difference between the 'Relation-only' and each of the 'Feature-only', and 'Relation + feature' conditions, $t(36)= 17.026$, $p < .001$; $t(36) = 18.982$, $p <$

.001, respectively. There was not a statistically significant difference between the 'Feature-only' and 'Relation + feature' conditions, $t(36)= -1.4446$, $p = 0.1572$.

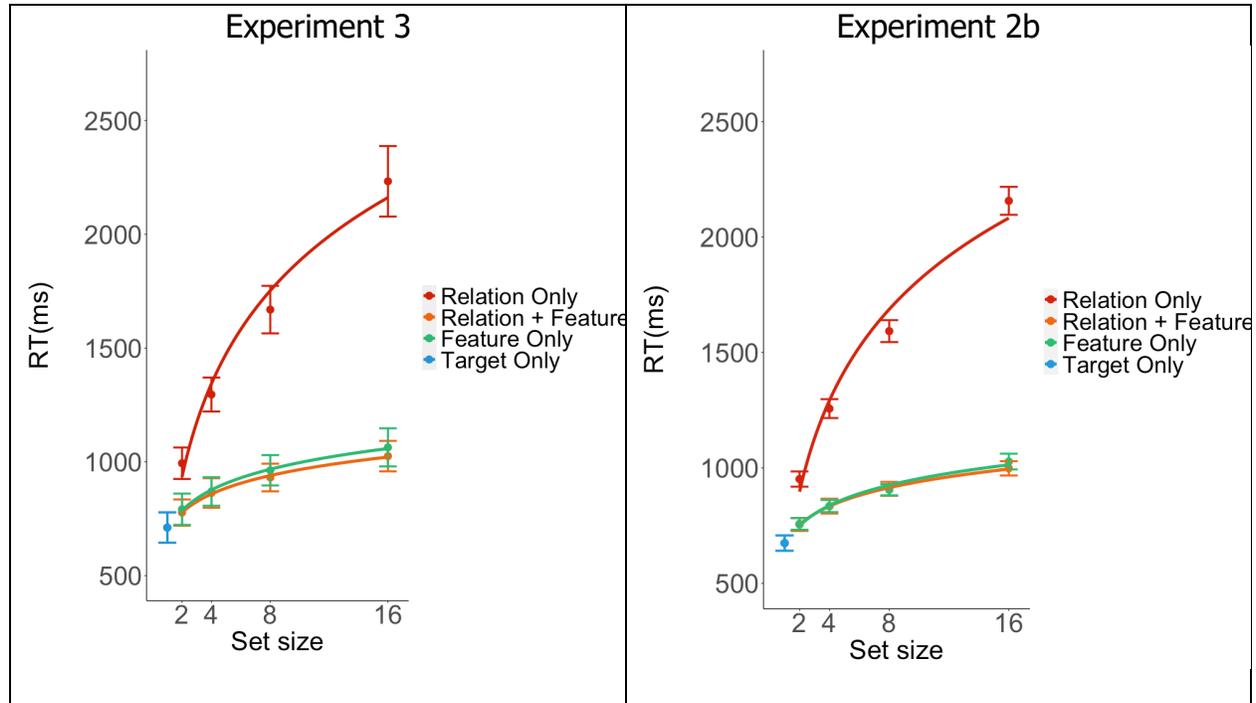

*Figure 5.6 Response time as a function of set size for Experiment 3 (left) and Experiment 2b (right). The 'Relation-only' condition is shown in red and the 'Feature-only' and 'Relation + feature' conditions are shown in green and orange, respectively. The 'Target Only' condition is shown in blue. The 'Relation-only' condition in both experiments had a steeper log slope than the 'Feature-only' and 'Relation + feature' conditions. For Experiment 3, there was not a statistically significant difference between the 'Feature-only' and 'Relation + feature' conditions. The data from Experiment 3 do not show a qualitative difference from the data in Experiment 2b, suggesting that the search items are not being processed holistically by special brain areas attuned to body parts. Error bars indicate the standard error of the mean. For Experiment 3, the mean log slopes were 536ms/log unit, 126ms/log unit, and 112ms/log unit for the 'Relation-only', 'Feature-only', and 'Relation + feature' conditions, respectively. The mean log slopes for Experiment 2b were 520ms/log unit, 123ms/log unit, and 115ms/log unit for the 'Relation-only', 'Feature-only', and 'Relation + feature' conditions, respectively.*

*Discussion*

Experiment 3 was designed to investigate whether the relational stimuli in Experiments 1a, 1b, 2a, and 2b were being processed holistically because of their potential resemblance to people. However, there were no qualitative differences between the results of Experiment 3 and the previous experiments, suggesting that parallel processing of these stimuli was not due to special holistic processing by brain areas attuned to body parts.

**Experiments 4a & 4b**

Can grouping by proximity facilitate a perceptual grouping process that makes it easier to detect a target in a relational search? If one examines a crowded search display from Experiments 1a or 1b, the target item creates a discontinuity that can be detected in the contours formed by the distractor items (Figure 5.7). Baker, Garrigan, & Kellman (2021) showed that constant curvature is important for contour grouping and segmentation. Bounded contours are quickly detected by the visual system (Kovacs & Julesz, 1993) and could help organize a search display for easier visual processing.

In Experiments 4a and 4b, items were spaced out to prevent the opportunity for such grouping by proximity to occur. The largest set size was eliminated, and steps were taken to ensure that items were spaced out in the radial array.

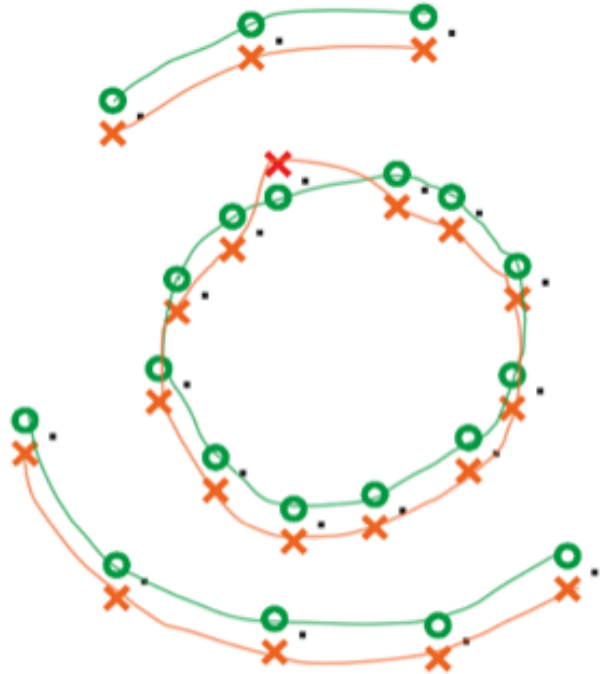

*Figure 5.7 A crowded search display from the 'Relation + feature' condition in Experiments 1a & 1b. The orange 'X' shapes form a contour, and the green 'O' shapes form a second contour. However, the target's red 'X' in the 'above' relation does not conform to the same curvature of the contours formed by the distractors.*

Methods

In Experiments 4a and 4b, the target and distractors were the same as in Experiments 1a and 1b, but the spacing of the search items was changed to reduce crowding. The total set sizes used were 1, 2, 4, and 8, with set size 16 omitted. To ensure an even distribution of search items in the display, the items were randomly assigned into one of four quadrants of the display. Additionally, the items were not allowed to be placed in adjacent locations within a circle, or in locations next to each other between the inner and outer circle. Aside from these constraints, the locations of the items were chosen randomly. Example displays from each condition in Experiments 4a and 4b are shown in Figure 5.8.

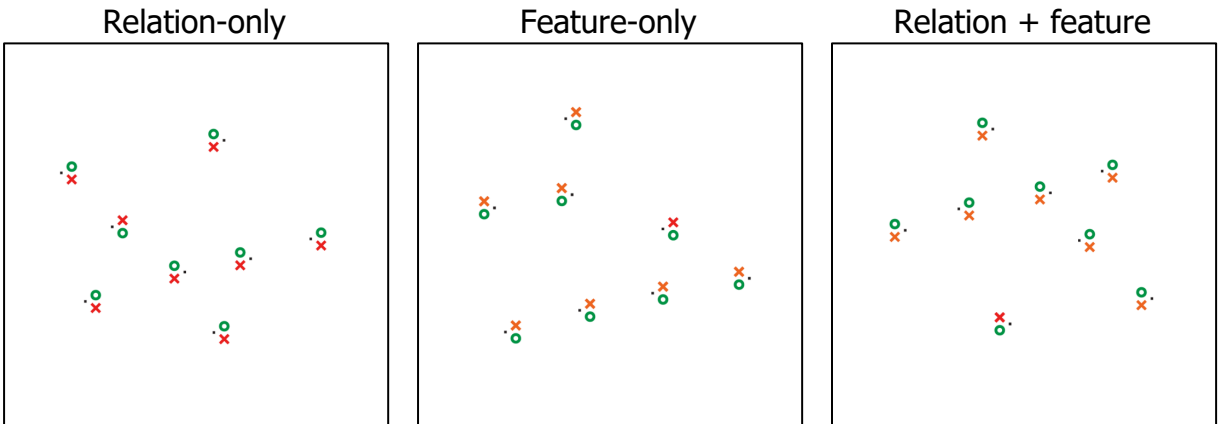

*Figure 5.8 Example displays for set size 16 for Experiment 3*

In Experiment 4a, in the 'Relation + feature' condition, black dots on the distractor items were only presented on one side. In Experiment 4b, this error was corrected, which served to both eliminate the possibility of an accidental confound and replicate the results of Experiment 4a.

The experimental and recruitment procedures were approved by the Institutional Review Board at the University of Illinois Urbana-Champaign. For Experiment 4a, 55 new participants were recruited from the Department of Psychology Subject Pool at the University of Illinois Urbana-Champaign during November 2020. Twenty-two participants were removed due to low accuracy (less than 90%), resulting in 33 included participants aged between 18 and 22 years old. For Experiment 4b, 42 new participants were recruited from the Department of Psychology Subject Pool at the University of Illinois Urbana-Champaign during September 2021. Six participants were removed due to low accuracy, resulting in the inclusion of 36 included participants aged between 18 and 21 years old.

*Results*

The results of Experiments 4a and 4b are shown in Figure 5.9. All conditions produced a negatively accelerating function of response time (RT) vs set size. A log slope was estimated per condition for each participant using a regression on log-transformed data, and the mean was computed across participants. For both experiments, the log slope for the 'Relation-only' condition was steeper than the log slopes for the 'Feature-only' and 'Relation + feature' conditions.

For Experiment 4a, the mean log slopes were 422ms/log unit, 201ms/log unit, and 208ms/log unit for the 'Relation-only', 'Feature-only', and 'Relation + feature' conditions, respectively. The estimated log slope per condition for each participant was used to perform paired t-tests between all combinations of conditions. The estimated log slope per condition was obtained for each participant, and paired t-tests were performed between all combinations of conditions. There was a statistically significant difference between the 'Relation-only' condition and each of the 'Feature-only' and 'Relation + feature' conditions, $t(32) = 11.209$, $p < .001$; $t(32) = 9.6116$, $p < .001$, respectively. There was not a statistically significant difference between the 'Feature-only' and 'Relation + feature' conditions, $t(32) = 1.7888$, $p = .0831$.

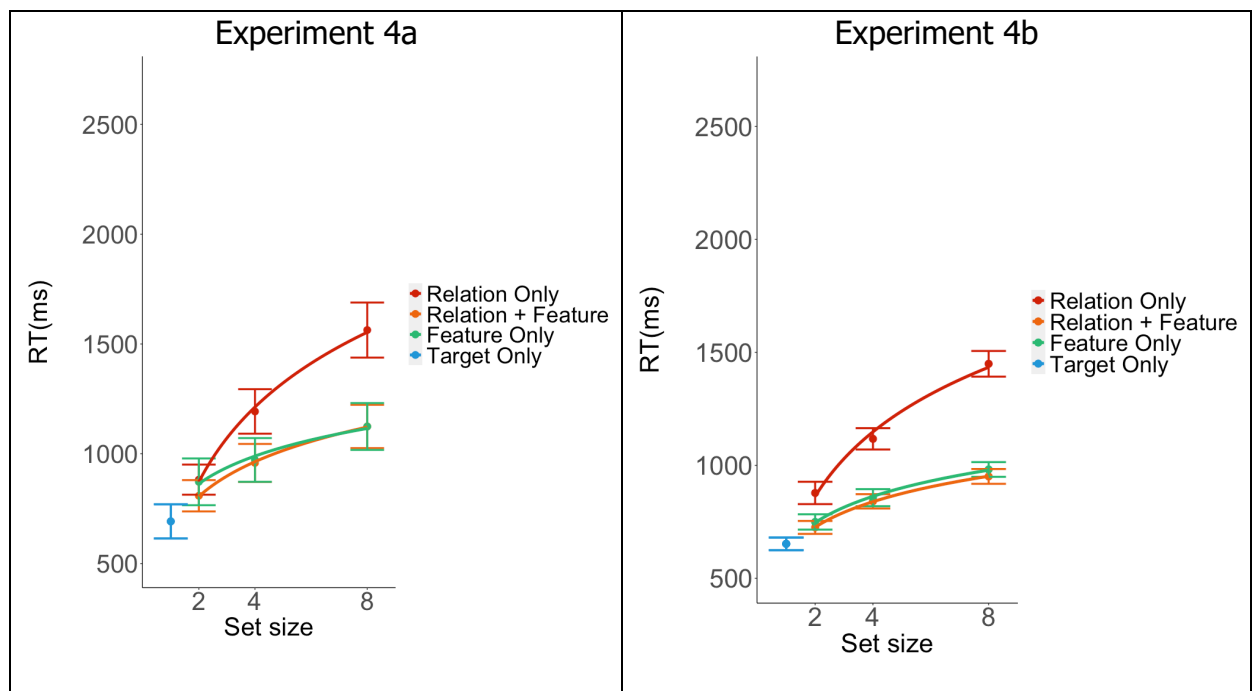

*Figure 5.9 Response time as a function of set size for Experiments 4a (left) and 4b (right). The 'Relation-only' condition is shown in red and the 'Feature-only' and 'Relation + feature' conditions are shown in green and orange, respectively. The 'Target Only' condition is shown in blue. The 'Relation-only' condition had a steeper log slope than the 'Feature-only' and 'Relation + feature' conditions. Unlike Experiments 1a and 1b, there was not a statistically significant difference between the 'Feature-only' and 'Relation + feature' conditions. For Experiment 4a, the mean log slopes were 422ms/log unit, 201ms/log unit, and 208ms/log unit for the 'Relation-only', 'Feature-only', and 'Relation + feature' conditions, respectively. For Experiment 4b, the mean log slopes were 380ms/log unit, 158ms/log unit, and 146ms/log unit for the 'Relation-only', 'Feature-only', and 'Relation + feature' conditions, respectively. The log slopes for the 'Relation-only' condition in Experiments 4a and 4b were shallower than the log slopes for the same condition in Experiments 1a and 1b. Error bars indicate the standard error of the mean.*

For Experiment 4b, the same analysis was performed as in Experiment 4a. The mean log slopes were 380ms/log unit, 158ms/log unit, and 146ms/log unit for the 'Relation-only', 'Feature-only', and 'Relation + feature' conditions, respectively. There was a statistically significant difference between the 'Relation-only' condition and each of the 'Feature-only' and 'Relation + feature' conditions, $t(35) = 9.3764$, $p < .001$; $t(35) = 12.957$, $p < .001$, respectively. There was not a statistically significant difference between the 'Feature-only' and 'Relation + feature' conditions, $t(35) = -0.24985$, $p =.8042$.

*Discussion*

Experiment 4b replicated the general pattern of data in Experiment 4a. In both experiments, all conditions again resulted in a negatively accelerating relationship between response time and set size, indicating that there was still parallel processing occurring. Unlike Experiments 1a and 1b, there was not a statistically significant difference between the 'Feature-only' and the 'Relation + feature' condition. The 'Relation-only' condition once again had a much larger log slope than either the 'Feature-only' condition or the 'Relation + feature' condition. However, the log slope was substantially lower than in Experiments 1a and 1b, which suggests that perceptual grouping was not facilitating parallel processing of the search displays.

**Summary**

The results of these seven experiments do not replicate the result of Logan's (1994) Experiment 1, which showed a linear relationship of response time to set size, indicating a fully serial search process. Logan used smaller set sizes, so there may have been limited opportunity to observe negative acceleration in response time as a function of set size.

The simulations in the CASPER model, which matched Logan's conclusions, also do not match the data in these experiments. Hence, the CASPER model is partially falsified for relational search displays like the ones in Experiments 1a-4b.

# Chapter 6: Simulating relational searches with emergent features

There are a few important differences between CASPER's predictions and the results of Experiments 1a-4b. First, although CASPER predicts a steep linear search function for the Relation-only condition in all experiments, the empirical data seem to produce a negatively accelerating search function. Indeed, the coefficient of determination for a log-transformed model is better than a basic linear model in all experiments. See Table 6.1 for comparisons.

| Experiment | Linear model | Log-transformed model |
|---|---|---|
| **1b** | $R^2 = 0.9593$ | $R^2 = 0.9730$ |
| **2b** | $R^2 = 0.9526$ | $R^2 = 0.9769$ |
| **3** | $R^2 = 0.9531$ | $R^2 = 0.9776$ |
| **4b** | $R^2 = 0.9631$ | $R^2 = 0.9909$ |

*Table 6.1 Coefficients of determination for linear vs. log-transformed statistical models of response time vs. set size for Experiments 1b, 2b, 3, and 4b. In all cases, the log-transformed model is a better fit than the linear model.*

A related issue with CASPER's predictions is that it does not capture the combined effect of Relation + feature information as found in Experiments 1a and 1b. It is possible that this problem would be resolved if CASPER captured parallel processing for relational information.

A third problem is that CASPER vastly underpredicts the difficulty of the Feature-only and relation+feature conditions relative to the Relation-only condition in Experiments 1b and 2b. A greater estimate of milliseconds per simulation iteration for any condition as compared to other conditions means that the model underestimates the difficulty of that condition. CASPER predicts 11.02ms/iteration and 8.97ms/iteration for the Feature-only and relation+feature conditions in Experiment 1b, respectively, and 19.82ms/iteration and 15.59ms/iteration for the same conditions in Experiment 2b. The Relation-only conditions are consistent with the estimates based on Logan's (1994) experiments, at 6.943ms/iteration and 6.06ms/iteration for Experiments 1b and 2b, respectively. For Simulations 1-6, all conditions ranged between 0.26 and 7.07ms/iteration.

**Possible causes for CASPER's incorrect predictions**

There are at least two possible reasons why CASPER's predictions don't match the empirical results presented here. The first possibility is that CASPER is wrong in its assumption that parallel processing doesn't happen over relations. However, true parallel processing of relations would be surprising considering the evidence suggesting that relational representations—and binding more generally—are degraded without foveal attention (Treisman, 1996; Rosenholtz, 2012).

A second possible reason for CASPER's failure to capture the results of Experiments 1a-4b is that the searches in the experiments are not really relational, but instead the visual system is able to extract some non-relational information from the stimuli that can be processed in parallel (e.g., with some sort of feature-based 'template'). One weakness of the stimuli used in the experiments presented here is that the Xs and Os are arranged in a constant location and size relative to one another across search items. This constancy makes it possible, at least in principle, to create a template or otherwise find an emergent feature that weakly distinguishes the target from distractors without using the relation. Such a template could be formed by weakly connected arrangements of common higher-order shapes (e.g., triangles or other surfaces), perhaps defined over negative space between the Xs and Os. For example, the negative space between the X and the O in the target 'points' upward, while the negative spaces in the distractors point downward. If the visual system is sensitive to this feature, even implicitly, then it could facilitate target-distractor discrimination during parallel processing.

**Simulation 8: Relations with multiple colors and an emergent feature**

Simulation 8 investigated whether CASPER can capture the negative acceleration of the search functions obtained in Experiments 1a-4b, proceeding under the tentative hypothesis that the visual properties of the search items in the experiments allow the visual system to obviate relational processing—perhaps by building a template for the search target as a whole, or by exploiting an emergent feature.

*Methods*

Recall that in Simulation 5, which simulated Pomerantz et al.'s (1977) effect of configuration on search, the addition of higher-order representations caused CASPER to predict improved search efficiency in the Good configural context condition relative to the Poor condition. This effect was mediated by the representational weight of the higher-order representation compared to the low-level representations. In Simulation 8, mid-level emergent features were added to CASPER's representations of the search items, and the salience of those features was manipulated to estimate the representational weight of that feature relative to the other features and relations. These new representations were encoded as one unit corresponding to the emergent feature for the target (red X above green O) and one unit corresponding to the emergent feature in the distractors (green O above red X). Otherwise, all parameters were the same as in previous simulations.

*Results*

The results of simulation 8 are shown in Figure 6.1. With the default salience ($\eta$ = 1.0), the emergent feature had too great an effect on the two relation-containing conditions, even driving the Relation-only search to greater efficiency than the Feature-only search (Figure 6.1B). However, when the salience for the emergent features was reduced to 1/3 of the default value ($\eta$ = 0.33), CASPER produced an excellent qualitative match to the results of Experiment 1b, with a negatively accelerating function of set size for all three conditions, including the Relation-only condition (Figure 6.1B). The efficiency of the relation+feature condition was also increased, driving the response time function lower than that of the Feature-only condition.

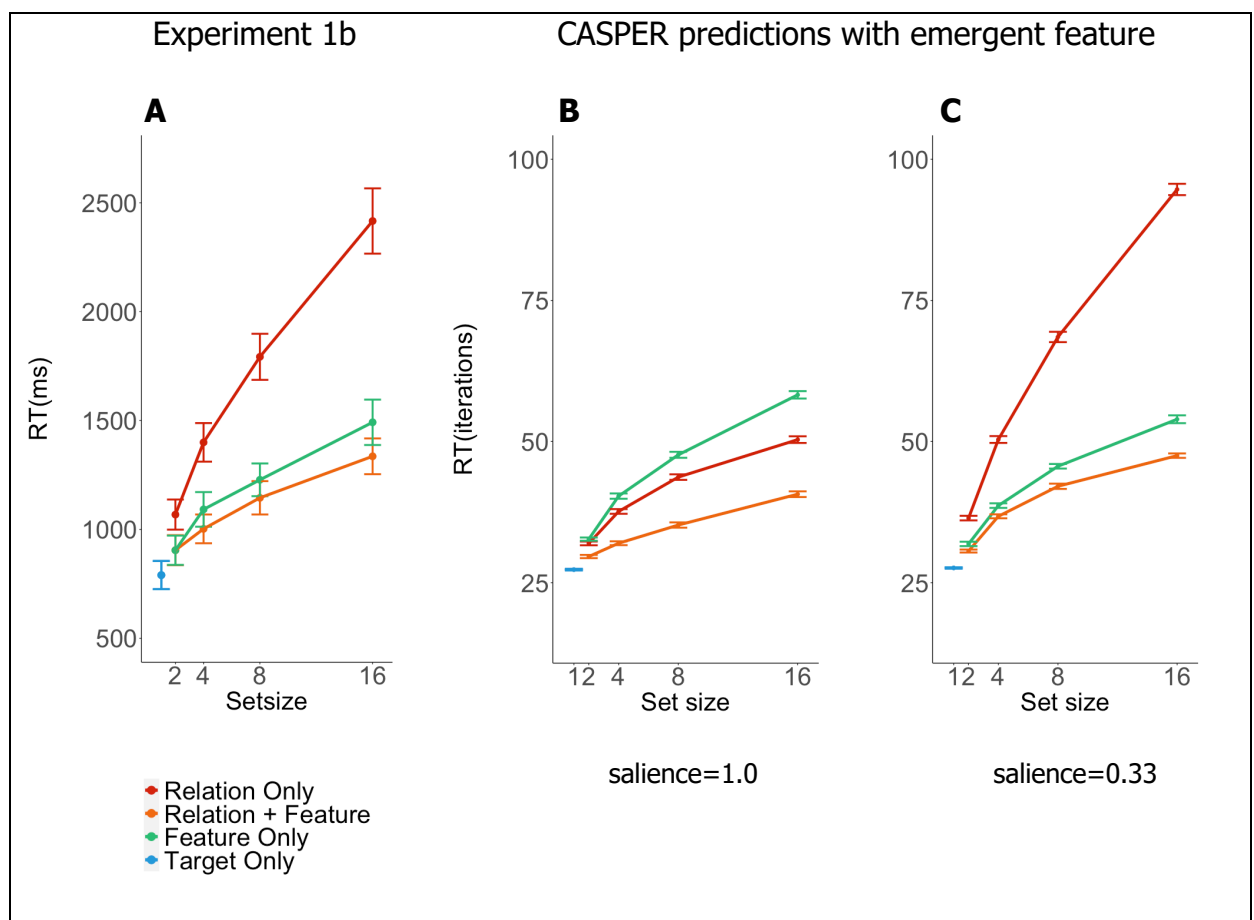

*Figure 6.1 A) Experiment 1b results. B & C) CASPER's simulations using the emergent feature at two different levels of salience, $\eta = 1$, and $\eta = 0.33$. The model captures the empirical results best when the emergent feature is assumed to have a lower level of salience relative to other features.*

The coefficients of determination for the simulation with salience $\eta = 0.33$ were $R^2 = 0.9996$ for the Relation-only condition, $R^2 = 0.9704$ for the Relation + feature condition, and $R^2 = 0.9921$ for the Feature-only condition. The combined coefficient for the model across all conditions was computed as in previous simulations, with $R^2 = 0.9959$. CASPER produced similar estimates of RT milliseconds per iteration across all conditions, at 20.1ms/iteration, 21.7ms/iteration, and 22.3ms/iteration for the Relation-only, relation+feature, and Feature-only conditions, respectively.

*Discussion*

The results of Simulation 8 suggest that the results of Experiments 1a-4b can be explained by the presence of an emergent feature in the search stimuli that can be used to weakly distinguish the target from the distractors. The salience of this emergent feature, however, must be assumed to be low relative to the other features in the search items for CASPER to appropriately characterize the results of Experiment 1b.

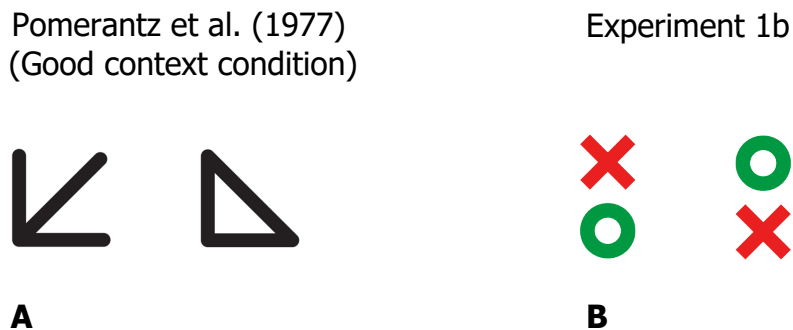

*Figure 6.2 A) Stimuli like those used by Pomerantz et al. (1977) in which the components of the stimuli are physically connected, one color, and arranged to form a familiar emergent feature. B) Experiment 1a and 1b stimuli, in which the emergent feature is hypothesized to be composed of physically separated components of differing colors that are not arranged to form a familiar feature (i.e., a weakly activated negative space that "points" up or down).*

Why would the emergent feature in Simulation 8 have low salience relative to other features, especially considering that the higher-order emergent features in Simulation 5, which simulated Pomerantz et al. (1977), seemed to have 128x the default representational strength of other features? One way to explain this difference is by considering the familiarity of the emergent features in each simulation. In Simulation 5, the higher-order emergent features are well-known familiar categories (i.e., arrows and triangles). In Simulation 8, although the hypothesized emergent surface features are metrically consistent across trials, the features are not familiar members of a common category (i.e., something one could name). Additionally, in Simulation 5, the components of the emergent features are physically connected and of a uniform color (Figure 6.2A). In Simulation 8, the components of the hypothesized emergent features are not physically connected and are different colors, potentially making perceptual grouping of a surface-like feature more difficult (Figure 6.2B).

These simulation results demonstrate that an emergent feature is a plausible account of the results from Experiments 1a and 1b, where the Relation-only conditions produced negatively accelerating search functions, and the relation+feature conditions were more efficient than Feature-only conditions.

**Simulation 9: Single color relations with an emergent feature**

Simulation 9 investigated whether the emergent features that seem to explain Experiment 1b, in which each search item was two different colors, also explain Experiment 2b, in which each search item was only one color. Recall that the results of Experiment 2b showed no additional benefit of relational information over feature information when color was diagnostic.

*Methods*

As in Simulation 8, mid-level emergent features were added to CASPER's representations of the search items. In Simulation 9, the emergent features were encoded as one unit corresponding the emergent feature for the target (red X above red O) and one unit corresponding to the emergent feature in the distractors (red O above red X). Because each search item in Experiment 2b was only one color, the salience of the emergent feature was returned to the default value of $\eta = 1.0$ in this simulation. Otherwise, all parameters were the same as in previous simulations.

On the hypothesis that the visual system doesn't exploit relational information when color information is diagnostic, a second simulation was run in which the emergent feature was removed from the Feature-only and relation+feature conditions but retained at salience $\eta = 1.0$ for the Relation-only condition.

*Results*

The results of Simulation 9 are shown in Figure 6.3 A and B. In Simulation 9, the fit to the Relation-only condition was excellent when the salience of the emergent feature was the same as the other features. However, the addition of this feature disrupted CASPER's predictions for the Feature-only and relation+ feature conditions, driving the relation+feature condition to greater efficiency than the Feature-only condition (Figure 6.3B). This difference is inconsistent with the results of Experiment 2b (Figure 6.3A), where the color feature information appeared to dominate search in the conditions in which it was available, leading to no additional efficiency from relational information. When the emergent feature is removed from the representations for these two conditions (but retained for the Feature-only condition) the model predicts similar search efficiency for the relation+feature and the Feature-only conditions (Figure 6.3C).

The coefficients of determination when the emergent feature was only included in the Relation-only condition were $R^2 = 0.994$ for the Relation-only condition, $R^2 = 0.993$ for the Relation + feature condition, and $R^2 = 0.971$ for the Feature-only condition. The combined coefficient for the model across all conditions was computed as in previous simulations, with $R^2 = 0.984$. CASPER's estimates of RT milliseconds per iteration were 19.8ms/iteration, 15.6ms/iteration, and 20.7ms/iteration for the Relation-only, relation+feature, and Feature-only conditions, respectively.

*Discussion*

The results of Simulation 9 are consistent with Simulation 8 in that they support the idea that an emergent feature is responsible for parallel processing in the relational stimuli used in the experiments reported here. However, changing the color properties of the stimuli seems to change the salience of the emergent feature. In the Relation-only condition, in which all the items in the display were one color, the model provides a good fit to the data when the salience of the emergent feature is about the same as the other features. In contrast, in Simulation 8, where colors were different within an item, the model provided a better fit when the emergent feature was less salient. However, when color feature information is available to easily distinguish the target from the distractors, the results of Simulation 9 suggest that the emergent feature has little to no influence on search, and the representation might as well not exist.

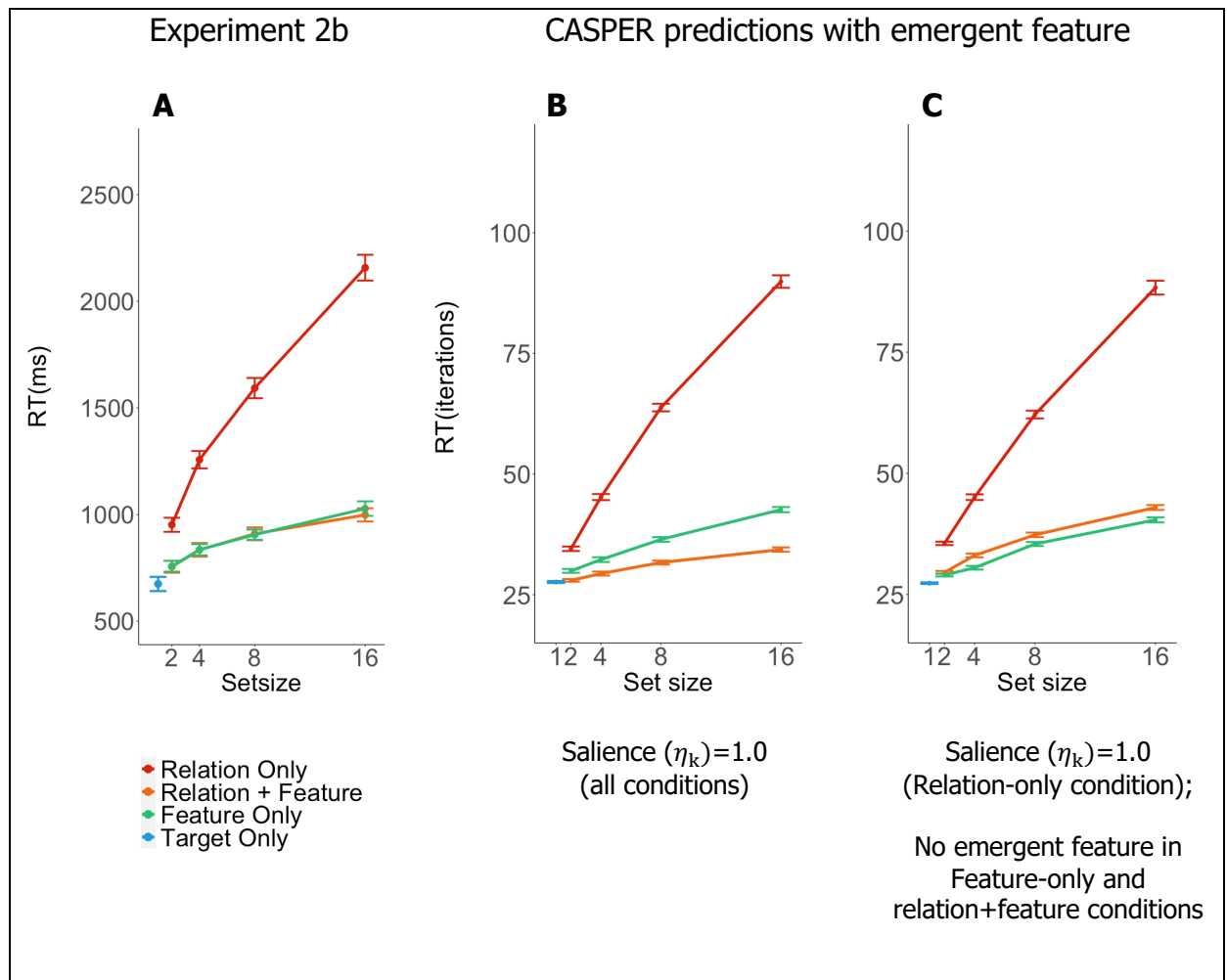

*Figure 6.3 A) Experiment 2b results. B) CASPER's simulation using the emergent feature when the salience of the feature $\eta_k = 1.0$. C) The model captures the empirical results best when the emergent feature is present in the Relation-only condition but is completely ignored in the other conditions.*

The results of Simulation 9 may also partially explain why Logan's (1994) Experiments differ from the results in Experiments 2a and 2b. Upon close examination, some negative acceleration in relational search functions could be present in Logan's experiments, but not to the degree as in the Relation-only conditions in Experiments 2a and 2b. Logan used fewer set sizes (up to 8 items), so the opportunity for negative acceleration to obtain was more limited than in the experiments presented here. However, it's also possible that any emergent features in Logan's stimuli were less salient. The search items in Experiments 2a and 2b contain rough approximations of surfaces in the negative space between the X and O. Logan's search items, which used a plus and a dash, do not have that property. If the visual system is looking for surfaces, then the presence of a surface-like emergent feature (even if only weakly connected) could be more salient (Elder & Zucker, 1993).

**Simulation 10: Increased efficiency in spaced out multicolor relational search**

CASPER accurately predicts the negatively accelerating search functions observed in Experiments 1b and 2b using a hypothesized emergent feature between the X and the O in the search items. In Experiments 4a and 4b, in which search items were spaced further apart, the Feature-only and relation+feature conditions produced very similar search functions. This is unlike Experiment 1b, where the relation+feature condition was more efficient than the Feature-only condition.

CASPER does not presently have a detailed account of the effect of spacing on feature and/or relation detection. However, it is possible that grouping by proximity and color makes it harder for human participants to perceive which Xs and Os go together to form a single item in a dense display during parallel processing: Palmer & Beck (2007) showed that participants are faster to identify target pairs of items when they have good grouping cues over color and proximity than when they have suboptimal grouping cues.

To approximate the effect of spacing out items so that they are easier to segment (and subsequently process), Simulation 10 modified the relevant sampling probabilities in CASPER. The logic behind this approach is that any difficulty segmenting items will increase the probability of sampling an irrelevant feature and/or decrease the probability of sampling a relevant feature. Although grouping may have a minimal effect on basic feature detection (e.g., X vs. O or red vs. green), it is likely to have a greater impact on the detection of emergent features formed between Xs and Os in the same item.

*Methods*

In Simulation 10, to simulate the effect of increased spacing between items, the probability of sampling relevant features was increased to 0.95 from the default value of 0.85. Otherwise, all methods were the same as in Simulation 8.

*Results & Discussion*

The results of Simulation 10 are shown in Figure 6.4B. CASPER shows a good qualitative fit to the Experiment 4b data (Figure 6.4A), with little difference between the Feature-only and relation+feature conditions. The coefficients of determination were $R^2$ = 0.9997 for the Relation-only condition, $R^2$ = 0.9978 for the Relation + feature condition, and $R^2$ = 0.9997 for the Feature-only condition. The combined coefficient for the model across all conditions was computed as in previous simulations, with $R^2$ =

0.9970. CASPER's estimates of milliseconds per iteration were 15.2ms/iteration, 17.4ms/iteration, and 15.2ms/iteration for the Relation-only, relation+feature, and Feature-only conditions, respectively.

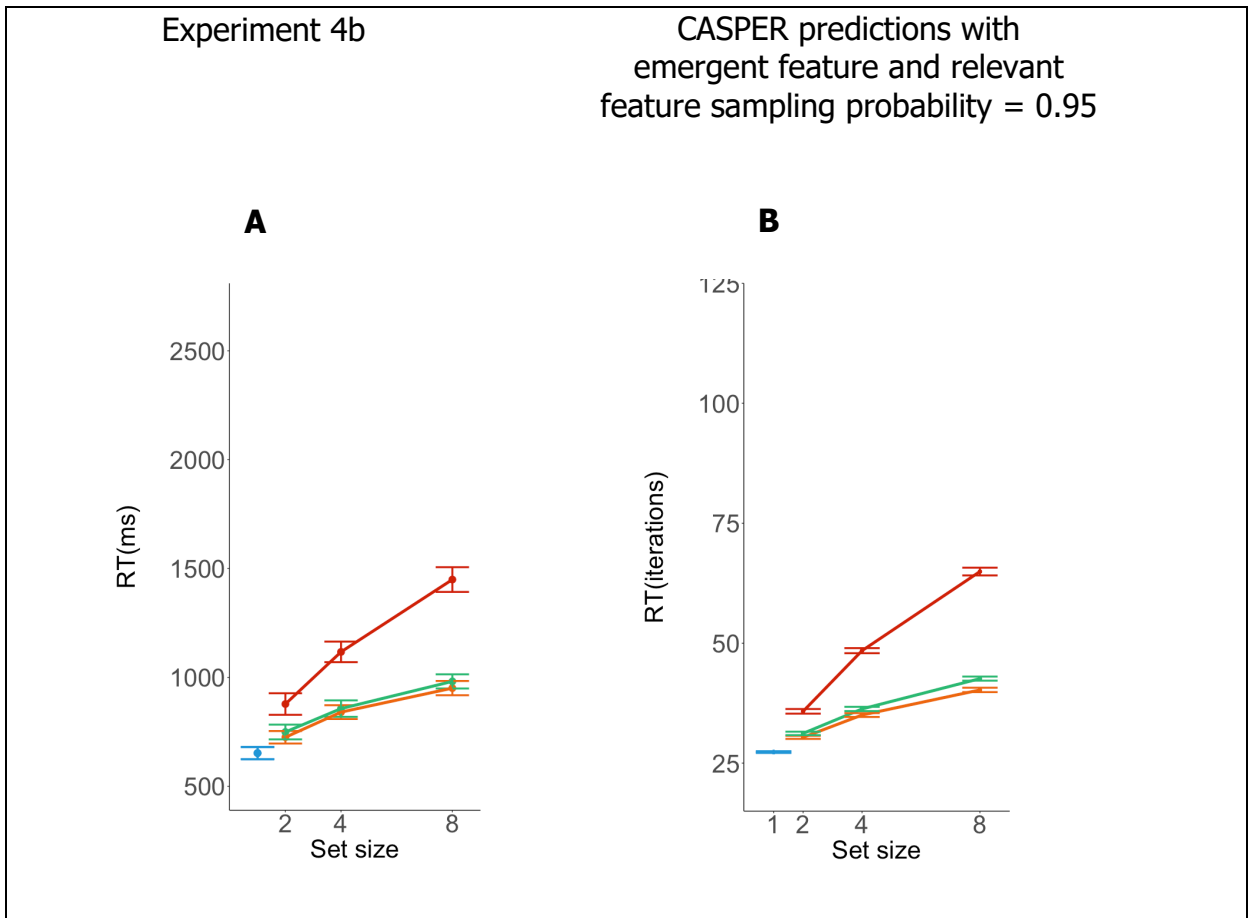

Figure 6.4 A) Experiment 4b results B) CASPER's simulation of Experiment 4b.

The results of Simulation 10 are consistent with the hypothesis that perceptual grouping by proximity and color can interfere with participants' ability to reliably sample features during parallel processing when grouping cues are suboptimal due to close spacing in the display.

**Summary**

Simulations 8-10 show that there is a possible confound in Experiments 1a-4b in which the visual system is using an emergent feature to weakly differentiate the target from distractors during parallel processing, leading to a weakly negatively accelerating search function in relational search. CASPER predicts that when within-item perceptual grouping cues are strong (as in Simulation 9, in which a search item was just one color) this emergent feature will be more salient, driving greater search efficiency. However, when color obviates the need to consider spatial relations, the emergent feature appears not to contribute to parallel processing (Experiments 2a and 2b and Simulation 9). When the within-item perceptual grouping cues are weaker (as in Simulations 8 and 10 where the items were composed of two different colored components) CASPER predicts the emergent feature will be less salient. CASPER also predicts (using sampling probability as a proxy for the difficulty in segmentation in lieu of a more detailed account of perceptual grouping) that close spacing of items will lead to less efficient search.

The possibility that an emergent feature is responsible for negatively accelerating search functions in putatively relational searches casts some doubt on the idea that relations can be processed in parallel. However, it remains possible that relations really are processed in parallel all over the visual field. At the present time, it's unclear whether Logan's (1994) relational experiments appeared to produce linear search functions because they did not have set sizes larger than 8 items that could have revealed a negative acceleration, or whether no emergent features in Logan's stimuli were salient enough to have an impact on the contribution of parallel processing in relational search.

## Chapter 7: General Discussion

CASPER is the first process model of visual search to explicitly incorporate a system of representation that the model uses to perform the search, and thus to make predictions. In Simulations 1 through 10, CASPER was able to simulate a range of search phenomena including efficient feature searches and inefficient conjunction searches (Treisman & Gelade, 1980), efficient conjunction searches (Wolfe et al., 1989), the effect of target-distractor similarity in feature searches (Buetti et al., 2016), search asymmetries (Treisman & Souther, 1985), configural effects (Pomerantz et al., 1977), and relational searches (Logan, 1994; Experiments 1a-4b).

CASPER produced good fits to all types of searches using its simplified representations of shape and color and one default set of tuning parameters. Slight modifications of salience were used to improve CASPER's fits to empirical data when the model's representations lacked the granularity to capture subtle differences. In the simulations reported here, the model produced reasonably consistent estimates of RT milliseconds per iteration within a simulation. However, across experiments the estimated milliseconds per iteration was greater for more difficult search tasks than for easier search tasks, meaning that CASPER tends to systematically underestimate the difficulty of difficult searches relative to easy ones.

The fact that CASPER makes it possible to calculate how many milliseconds in an experiment correspond to a single iteration in the model is one of its strengths. The fact that this estimate differs across different search tasks serves as an index of the difficulty of search (i.e., with larger ratios reflecting more difficult searches). More importantly, it reflects the limitations of the model due to its simplifying assumptions. One such simplification is the assumption that attentional scrutiny has a fixed temporal cost. This is clearly wrong. Nevertheless, it is a strength of the CASPER model that its operation is transparent enough to give us insights about where it is wrong and should be improved.

CASPER is useful a as model for understanding search and other visual phenomena precisely because its representations and processes are transparent, modular in the code, and designed specifically for the tasks they perform (e.g., the parallel match algorithm is designed specifically to match a target template to the various search items without the need for attentional intervention or dynamic feature binding). This property of CASPER differs from the dominant approach to modeling visual processing based on end-to-end differentiable neural networks, for which we have little insight into their internal operations (Bowers et al., 2023a, 2023b). CASPER is useful because we can understand how it works. Future work might include extensions to elucidate how CASPER's fundamental operations can be cached out in basic neural computations.

*Limitations of the model*

One of the clearest limitations of CASPER in its current state is that it ignores interactions among search items. As a result, it is unable to account for the effect of distractor-distractor similarity observed by Duncan & Humphreys (1989). As demonstrated in Simulations 1-10, many search phenomena can be simulated without an account of distractor-distractor similarity, but inter-item effects do appear to be relevant in some contexts. Search tasks in which there is no explicit target template, such as oddball tasks, are not currently part of CASPER's repertoire. Inter-item grouping effects would seem to be critical to any account of such tasks.

The most natural way to handle inter-item effects in a future version of CASPER would be to treat a group of similar distractors as a higher-order representational element (e.g., a surface), allowing the higher-order element to adjust the selection priorities of its lower-order constituents. This change would be consistent with Duncan & Humphreys' (1989) theory that posits rapid decay of similar distractors via perceptual grouping. Such a change to CASPER would require a substantially more sophisticated representational scheme than the model currently has, but the value in such an effort would be the ability to account for phenomena that seem to require processing that transcends item-based parallel processing (for an argument against item-based processing, see Hulleman & Olivers, 2017).

In the future CASPER could also tuned to handle target-absent searches. Wolfe & Chun (1996) have suggested that participants dynamically adjust the rate of parallel rejection during target-absent searches as a way to improve accuracy. No such dynamic adjustment exists in CASPER at the present time, but the model could be extended to include a mechanism that adjusts parameters based on error rates (e.g., due to learning over the course of the experiment).

**Parallel processing of relations**

Another avenue for future research concerns the negatively accelerating function of RT vs set size in the relational search experiments reported here. Recall that there are at least two broad ways to account for the negatively accelerating search function observed in Experiments 1a-4b: One is that relations, like features, really are detected in parallel all over the visual field and are being used like features for the rejection of distractor items. The other explanation is that a template or emergent feature could be guiding visual search.

CASPER's simulations of the negatively-accelerating shape of relational search results reported here are premised on the assumption that those experiments contain a confound. Specifically, the consistency of the shape of the search items gives rise to an emergent feature that the visual system exploits to differentiate the target from the distractors, even if only a little bit, during parallel processing. This is an assumption that needs to be tested with experiments. In particular, CASPER predicts that if a relational search paradigm can be created that contains no such diagnostic emergent feature, then search functions will be linear rather than logarithmic in the number of distractors.

In fact, there are two ways that search functions could be even worse than linear: if people make errors where they don't remember which items they searched, they might revisit them. The other way search functions could be worse than linear is if the target is defined not by the relations among its internal parts as in the experiments reported here, but by its relations to the distractors. In this case, search times would be expected to scale with the number of relations among items in the display, which for n items scales as $(n^2 - n)/2$ (Clevenger & Hummel, 2014).

In future work, additional experiments could be designed with the goal of presenting subjects with a more purely relational visual search task (one that does not permit the use of templates or emergent features, as elaborated below). If such experiments consistently demonstrate that visual relations can indeed be processed in parallel, then CASPER's eight core theoretical claim (that relational processing requires attention) will be falsified, and CASPER's parallel processing routines will have to be updated accordingly. This update will entail including relations in the search items and changing the parallel match algorithm to be more sensitive to the bindings of relational roles to fillers during parallel processing.

Experiments that provide evidence of parallel relational processing could be a challenge to design for two reasons. First, it is difficult to design relational stimuli without featural confounds that could shortcut the need for relational processing because any configuration of parts in relation to each other creates such emergent features. As long as these emergent features are reliably present across search items, the visual system might use them to perform the search task, resulting in a negatively accelerating search slope. Therefore, any task that produces a negative acceleration in the response time function could be called into question as to whether the task was truly relational.

The second difficulty is that removing featural confounds tends to make the relevant relations increasingly abstract (i.e., "find the red dot in the display that is closer to a green dot than any non-green dots", or "find the pair of letters that proceeds in a reverse alphabetical order among distractor pairs in forward alphabetical order"). Although this property is advantageous from the perspective of studying relational perception (rather than the perception of featural stand-ins for relations), the downside of this kind of control is that, as tasks become progressively more relational, they also start to resemble "cognitive" tasks.

The difficulty distinguishing perception from cognition in tasks involving visual relations is perhaps not surprising, given that we reason fluidly about abstract relationships between things that we see as a matter of daily life (e.g., "The beaver dam will prevent a floating object upstream from continuing downstream when the object arrives at the dam"). Indeed, the traditional division between perception and cognition may be more an artifact of scientists' need to divide the mind-brain into smaller comprehensible pieces than an actual reflection of its functional architecture.

As such, the quest for a purely perceptual relational search task may be folly. This problem does not imply that the visual search paradigm is useless for investigating relational processing, but it does mean that one must be careful about what one believes they are studying. This is where a model such as CASPER can be useful: if we can make a model account for a result in a putatively relational task using featural shortcuts, then that would suggest that the task may not be relational after all. Such an insight could play a crucial role in helping us to decide what experiments to run next.

# References

Baker, N., Garrigan, P., & Kellman, P. J. (2021). Constant curvature segments as building blocks of 2D shape representation. *Journal of Experimental Psychology. General, 150*(8), 1556–1580.

Biederman, I., & Cooper, E. E. (1991a). Priming contour-deleted images: Evidence for intermediate representations in visual object recognition. *Cognitive psychology, 23*(3), 393-419.

Biederman, I., & Cooper, E. E. (1991b). Evidence for Complete Translational and Reflectional Invariance in Visual Object Priming. *Perception, 20*(5), 585–593.

Bowers, J. S., Malhotra, G., Dujmović, M., Montero, M. L., Tsvetkov, C., Biscione, V., Puebla, G., Adolfi, F., Hummel, J.E., Heaton, R.F., Evans, B.D., Mitchell, J., & Blything, R. (2022). Deep problems with neural network models of human vision. *Behavioral and Brain Sciences*, 1-74.

Bravo, M. J., & Nakayama, K. (1992). The role of attention in different visual-search tasks. *Perception & Psychophysics, 51*(5), 465-472.

Buetti, S., Cronin, D. A., Madison, A. M., Wang, Z., & Lleras, A. (2016). Towards a better understanding of parallel visual processing in human vision: Evidence for exhaustive analysis of visual information. *Journal of Experimental Psychology: General, 145*(6), 672–707.

Bundesen, C. (1990). A theory of visual attention. *Psychological Review, 97*(4), 523–547.

Bundesen, C. (1998). A computational theory of visual attention. *Philosophical Transactions of the Royal Society of London. Series B: Biological Sciences, 353*(1373), 1271-1281.

Christoff, K., Prabhakaran, V., Dorfman, J., Zhao, Z., Kroger, J. K., Holyoak, K. J., & Gabrieli, J. D. (2001). Rostrolateral prefrontal cortex involvement in relational integration during reasoning. *Neuroimage, 14*(5), 1136-1149.

Chun, M. M., & Wolfe, J. M. (1996). Just say no: How are visual searches terminated when there is no target present? *Cognitive Psychology, 30*(1), 39-78.

Clevenger, P. E. (2017). *A role for long term memory in search for spatial relations* [Doctoral dissertation, University of Illinois at Urbana-Champaign].

Clevenger, P. E., & Hummel, J. E. (2014). Working memory for relations among objects. *Attention, Perception, & Psychophysics*, 76, 1933-1953.

Downing, P. E., Jiang, Y., Shuman, M., & Kanwisher, N. (2001). A cortical area selective for visual processing of the human body. *Science*, *293*(5539), 2470-2473.

Elder, J., & Zucker, S. (1993). The effect of contour closure on the rapid discrimination of two-dimensional shapes. *Vision Research*, *33*(7), 981-991.

Enns, J. T., & Rensink, R. A. (1991). Preattentive recovery of three-dimensional orientation from line drawings. Psychological Review, 98(3), 335–351.

Goodale, M. A., Milner, A. D., Jakobson, L. S., & Carey, D. P. (1991). A neurological dissociation between perceiving objects and grasping them. *Nature*, *349*, 154-156.

Green, C., & Hummel, J. E. (2006). Familiar interacting object pairs are perceptually grouped. Journal of Experimental Psychology: Human Perception and Performance, 32(5), 1107.

Hoffman, J. E. (1979). A two-stage model of visual search. *Perception & Psychophysics*, *25*(4), 319-327.

Hulleman, J., & Olivers, C. (2017). The impending demise of the item in visual search. *Behavioral and Brain Sciences*, *40*, E132.

Hummel, J. E., & Biederman, I. (1992). Dynamic binding in a neural network for shape recognition. *Psychological Review*, *99*(3), 480–517.

Kim, J. G., & Biederman, I. (2011). Where do objects become scenes? *Cerebral Cortex*, *21*(8), 1738-1746.

Kovacs, I., & Julesz, B. (1993). A closed curve is much more than an incomplete one: effect of closure in figure-ground segmentation. *Proceedings of the National Academy of Sciences, 90*(16), 7495-7497.

Kristjánsson, Á. (2015). Reconsidering Visual Search. *I-Perception*, *6*(6).

Kroger, J. K., Sabb, F. W., Fales, C. L., Bookheimer, S. Y., Cohen, M. S., & Holyoak, K. J. (2002). Recruitment of anterior dorsolateral prefrontal cortex in human reasoning: a parametric study of relational complexity. *Cerebral Cortex*, *12*(5), 477-485.

Lleras, A., Rensink, R. A., & Enns, J. T. (2005). Rapid resumption of interrupted visual search: New insights on the interaction between vision and memory. *Psychological Science*, *16*(9), 684-688.

Lleras, A., Wang, Z., Ng, G. J. P., Ballew, K., Xu, J., & Buetti, S. (2020). A target contrast signal theory of parallel processing in goal-directed search. *Attention, Perception, & Psychophysics*, *82*, 394-425.

Logan, G. D. (1994). Spatial attention and the apprehension of spatial relations. *Journal of Experimental Psychology: Human Perception and Performance*, *20*(5), 1015–1036.

Logan, G. D. (1996). The CODE theory of visual attention: An integration of space-based and object-based attention. *Psychological Review*, *103*(4), 603–649.

Luce, R. D. (1959). On the possible psychophysical laws. *Psychological Review*, *66*(2), 81–95.

Markman, A. B., & Gentner, D. (1993). Structural alignment during similarity comparisons. *Cognitive Psychology*, *25*(4), 431-467.

Marr, D. (1982). *Vision*. San Francisco, CA: Freeman.

Müller, H. J., Heller, D., & Ziegler, J. (1995). Visual search for singleton feature targets within and across feature dimensions. *Perception & Psychophysics*, *57*(1), 1-17.

Nakayama, K., He, Z. J., & Shimojo, S. (1995). Visual surface representation: A critical link between lower-level and higher-level vision. In S. M. Kosslyn & D. N. Osherson (Eds.), *Visual cognition: An invitation to cognitive science* (pp. 1–70). The MIT Press.

Palmer, S. E. (1977). Hierarchical structure in perceptual representation. *Cognitive Psychology*, *9*(4), 441-474.

Palmer, S., & Rock, I. (1994). Rethinking perceptual organization: The role of uniform connectedness. *Psychonomic Bulletin & Review*, *1*(1), 29-55.

Pashler, H. (1990). Coordinate frame for symmetry detection and object recognition. *Journal of Experimental Psychology: Human Perception and Performance, 16*(1), 150–163.

Pomerantz, J. R., Sager, L. C., & Stoever, R. J. (1977). Perception of wholes and of their component parts: Some configural superiority effects. *Journal of Experimental Psychology: Human Perception and Performance, 3*(3), 422–435.

Rosenholtz, R., Huang, J., & Ehinger, K. A. (2012). Rethinking the role of top-down attention in vision: Effects attributable to a lossy representation in peripheral vision. *Frontiers in Psychology*, 3, 13.

Singh, M., & Hoffman, D. D. (2001). Part-based representations of visual shape and implications for visual cognition. In *Advances in psychology* (Vol. 130, pp. 401-459). North-Holland.

Stankiewicz, B. J., Hummel, J. E., & Cooper, E. E. (1998). The role of attention in priming for left–right reflections of object images: Evidence for a dual representation of object shape. *Journal of Experimental Psychology: Human Perception and Performance, 24*(3), 732–744.

Tanaka, K. (1996). Inferotemporal cortex and object vision. *Annual review of neuroscience*, *19*(1), 109-139.

Taylor, J. C., Wiggett, A. J., & Downing, P. E. (2007). Functional MRI analysis of body and body part representations in the extrastriate and fusiform body areas. *Journal of Neurophysiology*, *98*(3), 1626-1633.

Thoma, V., & Davidoff, J. (2002). Priming for depth-rotated objects depends on attention. *Journal of Vision*, *2*(7), 42-42.

Thoma, V., Davidoff, J., & Hummel, J. E. (2007). Priming of plane-rotated objects depends on attention and view familiarity. *Visual Cognition*, *15*(2), 179-210.

Thornton, T. L., & Gilden, D. L. (2007). Parallel and serial processes in visual search. *Psychological Review*, *114*(1), 71.

Tohill, J. M., & Holyoak, K. J. (2000). The impact of anxiety on analogical reasoning. Thinking & Reasoning, *6(*1), 27-40.

Townsend, J. T. (1990). Serial vs. parallel processing: Sometimes they look like Tweedledum and Tweedledee but they can (and should) be distinguished. *Psychological Science*, *1*(1), 46-54.

Treisman, A. (1996). The binding problem. *Current opinion in neurobiology*, *6*(2), 171-178.

Treisman, A. M., & Gelade, G. (1980). A feature-integration theory of attention. *Cognitive psychology*, *12*(1), 97-136.

Treisman, A., & Souther, J. (1985). Search asymmetry: A diagnostic for preattentive processing of separable features. *Journal of Experimental Psychology: General, 114*(3), 285–310.

Vannuscorps, G., Galaburda, A., & Caramazza, A. (2022). Shape-centered representations of bounded regions of space mediate the perception of objects. *Cognitive Neuropsychology, 39*(1-2), 1-50.

Wenderoth, P. (1994). The salience of vertical symmetry. *Perception, 23*(2), 221-236.

Wolfe, J. M. (1994). Guided search 2.0 a revised model of visual search. *Psychonomic bulletin & review, 1*, 202-238.

Wolfe, J. M., Cave, K. R., & Franzel, S. L. (1989). Guided search: An alternative to the feature integration model for visual search. *Journal of Experimental Psychology: Human Perception and Performance, 15*(3), 419–433.

Wolfe, J. M. (2007). Guided Search 4.0: Current progress with a model of visual search. In W. D. Gray (Ed.), *Integrated models of cognitive systems* (pp. 99–119). Oxford University Press.